\pdfoutput=1

\documentclass[11pt,table,xcdraw]{article}

\usepackage[final]{acl}

\usepackage{times}
\usepackage{latexsym}

\usepackage[T1]{fontenc}

\usepackage[utf8]{inputenc}
\usepackage{microtype}

\usepackage{inconsolata}

\usepackage{graphicx}

\usepackage[most]{tcolorbox} 
\tcbset{
casebox/.style={
enhanced,
colback=white, 
colframe=black, 
colbacktitle=gray!20, 
coltitle=black,
fonttitle=\bfseries, 
fontupper=\normalsize, 
width=\textwidth, 
boxrule=0.5mm, 
arc=3mm, 
left=1mm, right=1mm, top=1mm, bottom=1mm, 
boxsep=5pt, 
rounded corners, 
}
}

\tcbset{
promptbox/.style={
enhanced,
colback=white, 
colframe=black, 
colbacktitle=gray!20, 
coltitle=black,
fonttitle=\bfseries, 
fontupper=\normalsize, 
width=\textwidth, 
boxrule=0.5mm, 
arc=3mm, 
left=1mm, right=1mm, top=1mm, bottom=1mm, 
boxsep=5pt, 
rounded corners, 
}
}
\usepackage{CJKutf8}
\usepackage{xcolor} 
\usepackage{booktabs}
\usepackage{multirow}
\usepackage{amsmath}
\usepackage{colortbl}
\usepackage{adjustbox}
\usepackage[normalem]{ulem}
\usepackage{cuted}
\usepackage{afterpage}
\usepackage{amssymb}
\usepackage{float}
\usepackage{caption}
\usepackage{subcaption} 
\usepackage{enumitem} 
\usepackage{amssymb}

\definecolor{lightgray}{gray}{0.8}

\newcommand{\graycell}[1]{\cellcolor{lightgray!50}#1}

\title{CNNSum: Exploring Long-Context Summarization with\\Large Language Models in Chinese Novels}

\author{%
\textbf{Lingxiao~Wei\textsuperscript{$\dagger$}}
\textbf{\quad He~Yan\textsuperscript{$\text{\maltese}$}}
\textbf{\quad Xiangju~Lu\textsuperscript{$\text{\maltese}$}}
\textbf{\quad Junmin~Zhu\textsuperscript{$\text{\maltese}$}}\\
\textbf{\quad Jun~Wang}\textsuperscript{$\dagger$}
\textbf{\quad Wei~Zhang}\textsuperscript{$\dagger\ddagger$}\thanks{Corresponding author.}\\
\textsuperscript{$\dagger$}East China Normal University
\quad
\textsuperscript{$\text{\maltese}$}iQIYI Inc 
\quad
\textsuperscript{$\ddagger$}Shanghai Innovation Institute
\\
\texttt{51265901053@stu.ecnu.edu.cn} 
\quad 
\texttt{zhangwei.thu2011@gmail.com}\\
  }

\begin{document}
\begin{CJK}{UTF8}{gbsn} 
\maketitle
\begin{abstract}
Large language models (LLMs) have been well-researched in various long-context tasks. 
However, the scarcity of long-context summarization datasets hinders progress in this area.
To address this, we introduce CNNSum, a multi-scale long-context summarization benchmark based on Chinese novels, featuring human-driven annotations across four subsets totaling 695 samples, with lengths ranging from 16k to 128k.
We benchmark numerous LLMs and conduct detailed human assessments to summarize abnormal output types. Furthermore, we extensively explore how to improve long-context summarization.
In our study:
(1) Advanced LLMs may generate much subjective commentary, leading to vague summaries.
(2) Currently, long-context summarization mainly relies on memory ability. The advantages of Large LLMs are hard to utilize, thus small LLMs are more cost-effective.
(3) Different prompt types paired with various version models may cause large performance gaps. In further fine-tuning, these can be mitigated, and the Base version models perform better.
(4) LLMs with RoPE-base scaled exhibit strong extrapolation potential; using short-context data can significantly improve long-context summarization performance. However, further applying other interpolation methods requires careful selection.
(5) CNNSum provides more reliable evaluation results than other benchmarks. 
We release CNNSum to advance future research.\footnote{\url{https://github.com/CxsGhost/CNNSum}}
\end{abstract}

\section{Introduction}
Long-context large language models are advancing rapidly~\cite{wang2024beyondsurvey,pawar2024andsurvey}, making complex long-context tasks that were once considered difficult achieve revolutionary progress~\cite{li2024needlebench}.
However, long-context summarization tasks progress slowly, which is crucial for assessing long-context ability and forms the basis for complex contextual understanding~\cite{pagnoni2021understanding,bhandari2020re,kim2024fables}. 
~\cite{pu2023sumdied} declare that traditional summarization research is dead, but creating new high-quality datasets remains essential. 
Unlike most tasks, summarization is more subjective, without a gold standard.
And obtaining high-quality data is challenging for long texts, which requires a global context understanding and memory, and even human experts struggle to annotate directly~\cite{qiu2024clongeval}. 

Recent studies on LLMs summarization~\cite{tam2023evalfactual,song2024finesure,li2024slidesum} are still based on previous short datasets~\cite{narayan2018xsum,nallapati2016cnndm}.
~\cite{bertsch2024unlimiformer} proposes unlimited-input transformers based on summarization task, using only BookSum~\cite{kryscinski2021booksum} as sufficiently long data, which remains the latest book-length dataset. The maximum length of newer Chinese long summarization datasets~\cite{liu2022clts+} is only 6k.
~\cite{chang2023booookscore,kim2024fables} study book-length summarization, but both are expensive in annotating their own datasets to avoid leakage risks.

Most studies focus on hallucination-related issues evaluations. 
However, implementing these evaluations holds limited significance if models cannot work properly.
Although the 128k context length has been common, LLMs exhibit large performance degradation on various tasks as sequence length increases~\cite{hsieh2024ruler}. In this study, we indeed find that outputs may be disorganized, meaningless, or fail to follow the instructions. 
The core is how LLMs handle long-context summarization tasks. 
Due to a lack of datasets, there is still insufficient systematic research and guidance.

Despite many long-context LLMs benchmarks emerging~\cite{an2023leval,bai2023longbench,kwan2023m4le,zhang2024infty,ni2024xlbench,qiu2024clongeval}, their summarization tasks still have considerable room for improvement in terms of construction. Regardless of language and domain, we summarize the shortcomings: (1) Building on old datasets~\cite{fabbri2019multinews,huang2021govreport,sharma2019bigpatent,kryscinski2021booksum,wang2021cnewsum} leads to high leakage risks~\cite{xu2024benchmarkingleak,li2024openleak}. (2) Data volume is small, maybe just dozens. (3) The average and maximum lengths are short, typically under 16k. (4) Lacking multi-scale length-based subsets limits evaluation in varying contexts, and our experiments indicate a wide range of mixed lengths may lead to misleading results. (5) Annotations are either collected from the web, carrying high leakage risk and low quality, or synthesized by LLMs and may have various errors~\cite{chang2023booookscore,song2024finesure,kim2024fables}. Each benchmark mentioned above has at least two of these shortcomings.

To better explore long-context summarization with LLMs, we introduce Chinese Novel Summarization (CNNSum), a new multi-scale dataset that improves all the above shortcomings. Based on a newly collected corpus, we built four subsets: L(16k), XL(32k), 2XL(64k), and 3XL(128k), totaling 695 samples. The annotations are completed by human experts with assistance from LLMs. Since prompt formats greatly impact LLMs~\cite{sclar2024prompt} and beginning and ending tokens are crucial in long contexts~\cite{han2024lminfinite,liu2024lost}, we define two prompt types for baselines: Instruction at the Beginning (Prompt-IB) and Instruction at the End (Prompt-IE), as shown in Appendix~\ref{two_baseline_prompts}.
Although ROUGE-L~\cite{lin2004rouge} remains the primary metric in most benchmarks, it differs significantly from human preferences. Even LLMs are not yet reliable enough to replace humans~\cite{llmnotenough}. While referencing automatic metrics, we conduct fine-grained inspection and analysis of the outputs to categorize error types. 
Our observations are as follows: (1) GPT-4o~\cite{openai2024} prefers more subjective narration leading to vague summaries, so focusing on objective plots remains the key.
(2) The two prompts can result in a significant output quality gap across various LLMs. 
(3) The Chat or Instruction version may harm the Base model's long-context summarization ability. 
(4) The large LLMs struggle to fully leverage their advantages in reasoning and comprehension, while small LLMs are more cost-effective.

The abnormal behaviors are usually related to training configurations of LLMs; we further conduct extensive fine-tuning experiments. The LLMs may have acquired longer-context abilities during pre-training~\cite{xiong2023effective,fu2024data}, requiring only activation. Consequently, we concatenate short summarization data into longer ones for training.
Our findings indicate that:
(1) The performance gaps caused by prompts can be mitigated, allowing models to fully perform.
(2) The Base models are better for further fine-tuning and extrapolation than the Instruction or Chat version.
(3) Fine-tuning with short-context data can significantly improve long-context summarization performance.
(4) Current LLMs widely use Adjusted Base Frequency (ABF)~\cite{xiong2023effective}, improving the extrapolation potential and maintaining decent extrapolation performance on samples several times longer than the training length. However, when further combined with other interpolation methods, they show interesting differences.
(5) CNNSum could provide more reliable evaluations, as it employs a more rigorous and reasonable length sampling strategy. The only relatively new and large-scale Chinese summarization benchmark is CLongEval-LStSum~\cite{qiu2024clongeval}, which reuses BookSum~\cite{kryscinski2021booksum} and annotations are synthesized by GPT-4. However, we find it may produce misleading results compared to CNNSum, especially in extrapolation evaluation.

In summary, we introduce CNNSum, an entirely new long-context Chinese novel summarization benchmark, which is significantly superior in design and construction compared to existing datasets. We carry out comprehensive explorations of long-context summarization tasks for LLMs, providing guiding insights for future research in this field.

\section{Related Work}

\subsection{Summarization Evaluation of LLMs}
ROUGE~\cite{lin2004rouge} is still popular due to its simple implementation. It measures the information overlap between outputs and references.~\cite{chang2023booookscore} proposes a protocol for coherence evaluation of book-length summaries generated by LLMs and achieves an LLM-based automatic implementation. It does not focus on faithfulness and thus does not rely on gold summaries but advanced LLMs like GPT-4~\cite{achiam2023gpt4}.~\cite{tam2023evalfactual} constructs a benchmark for evaluating factual consistency by manually re-annotating previous news summarization datasets~\cite{nallapati2016cnndm}.~\cite{krishna2023longeval} proposes guidelines for human evaluation of faithfulness.~\cite{song2024finesure} introduces a prompt-guided fine-grained evaluator for faithfulness and completeness. It also relies on advanced LLMs for reliable evaluation.~\cite{sumcontextentire} studies how LLMs utilize entire context in summarization, still based on those previous datasets.~\cite{kim2024fables} is the first to conduct a large-scale evaluation of book-length summarization faithfulness, which mainly focuses on commercial models.~\cite{2024analyzingcodesum} analyzes code summarization performance of LLMs based on short-context datasets.

\subsection{Long-Context Extension for LLMs}

RoPE~\cite{su2024roformer} is widely used in LLMs, but its extrapolation performance is limited. Position Interpolation~\cite{chen2023pi} achieves efficient extension but still needs fine-tuning on the target length. NTK methods~\cite{reddit_ntkaware_scaled_rope,reddit24dyntk} mitigate the high-frequency information loss caused by interpolation, enabling extrapolation directly. YaRN~\cite{peng2023yarn} combines the above methods and attains 2x extrapolation. CLEX~\cite{chen2023clex} achieves more efficient dynamic scaling, and LongRoPE~\cite{ding2024longrope} leverages RoPE's non-uniformity to search better interpolations for more times of extrapolation, but the iteration process is complex. The above methods are mostly implemented on early LLMs~\cite{touvron2023llama,touvron2023llama2} for experiments. 
The Adjusted Base Frequency (ABF)~\cite{xiong2023effective} has been widely used~\cite{yang2024qwen2,qwen2.5,young2024yi,cai2024internlm2}, significantly improving extrapolation potential.~\cite{liu2023scalinglaws} further proposes scaling laws for RoPE base. There is still no systematic research on applying other interpolation methods to ABF-scaled LLMs.

\section{CNNSum Construction}
\label{bm_construction}
\subsection{Chinese Book Corpus Collection}
We collect 103 books from the Chinese Internet open-source data, each having a clear chapter structure. We remove books that are composed of multiple independent short stories or that lack main storylines, as their long-context dependencies are weak. Qwen2-72B-Instruct~\cite{yang2024qwen2} is used to detect potentially popular books, as it has a high leakage risk~\cite{ni2024trainingbench,xu2024benchmarkingleak}. We randomly select chapters from each book to generate summaries. If the model identifies source books and outputs extra information, we remove them as they may also be extensively leaked in other LLMs; this filters out 27 books.
Our corpus has many web-serialized novels, often exhibiting unique writing habits or styles of authors. Common cases include non-standard punctuation usage and randomly interspersed irrelevant extras, as shown in Appendix~\ref{sec:yuliao_case}. These may disrupt the context coherence or lead to unexpected behavior in model outputs. We format these via regular expressions and manual checks.

\begin{table*}[t]
  \centering
  \setlength{\tabcolsep}{0.25em}  
  \resizebox{\linewidth}{!}{
  \renewcommand{\arraystretch}{1.05}
  \begin{tabular}{l cccc | cccc | cccc | cccc}
    \hline
    \multirow{3}{*}{\centering \textbf{Model Series}} & \multicolumn{4}{c}{\textbf{CNNSum-L (Count=190)}} & \multicolumn{4}{c}{\textbf{CNNSum-XL (Count=195)}} & \multicolumn{4}{c}{\textbf{CNNSum-2XL (Count=200)}} & \multicolumn{4}{c}{\textbf{CNNSum-3XL (Count=110)}}\\
    & \multicolumn{4}{c}{\textbf{Source Books=76}} & \multicolumn{4}{c}{\textbf{Source Books=71}} & \multicolumn{4}{c}{\textbf{Source Books=60}} & \multicolumn{4}{c}{\textbf{Source Books=45}}\\
     & Min & Q1 & Mean & Max & Min & Q1 & Mean & Max & Min & Q1 & Mean & Max & Min & Q1 & Mean & Max\\  
    \hline
    Yi / Yi-1.5   & 13,447 & 14,841 & 15,721 & 18,323 & 27,586 & 30,932 & 31,844 & 34,703 & 56,323 & 63,683 & 64,303 & 67,403 & 115,351 & 128,673 & 128,751 & 132,688\\
    Llama3 / Llama3.1 & 15,128 & 18,203 & 19,097 & 23,408 & 31,795 & 37,344 & 38,557 & 42,659 & 67,952 & 76,326 & 77,845 & 85,690 & 140,920 & 152,603 & 155,339 & 166,504 \\
    Qwen1.5 / Qwen2.5   & 12,938 & 14,529 & 15,324 & 18,501 & 27,130 & 30,189 & 31,026 & 34,104 & 54,760 & 61,718 & 62,526 & 66,844 & 110,373 & 123,792 & 124,995 & 130,707 \\
    GPT-4o-2024-08-06   & 15,564 & 17,974 & 18,814 & 22,792 & 32,428 & 36,676 & 37,984 & 42,177 & 68,641 & 74,996 & 76,598 & 85,231 & 138,417 & 150,226 & 152,875 & 164,939 \\
    \hline
  \end{tabular}}
      \vspace{-0.5em}
  \caption{
    Token sequence length of CNNsum. Note that GPT-4o uses tiktoken\protect\footnotemark and Q1 denotes quarter quartile.}
\vspace{-1em}
\label{bm_info}
\end{table*}
\footnotetext{\url{https://github.com/openai/tiktoken}}

\subsection{Multi-Scale Sampling}
\label{sec:bm_length_settings}
We calculate the efficiency of various tokenizers in tokenizing Chinese text based on our corpus; details are in Appendix~\ref{sec:llms_cibiao}.
We use Yi~\cite{young2024yi} tokenizer to process chapters and define four token-level target lengths $T$ and their ranges. Each range has a specific lower boundary to avoid scoring bias from overly short data.
The upper boundary is uniformly set as $T+2k$ ($k=1024$), able to evaluate extrapolation without causing score collapse. 
The specific settings are as follows:
\begin{figure}[!h]
\vspace{-0.5em}
    \centering
\begin{adjustbox}{max width=\linewidth}
$
\begin{aligned}
T_{L} &= 16k, & Range_{L} &= [16k - 4k, 16k + 2k] \\
T_{XL} &= 32k, & Range_{XL} &= [32k - 6k, 32k + 2k] \\
T_{2XL} &= 64k, & Range_{2XL} &= [64k - 10k, 64k + 2k] \\
T_{3XL} &= 128k, & Range_{3XL} &= [128k - 16k, 128k + 2k]
\end{aligned}
$
\end{adjustbox}
\vspace{-1em}
\end{figure}

\noindent The book $B$ is composed of multiple chapters $c$, represented as $B=\{c_0, c_1, \ldots, c_i \}$.
For a target range $R$ and a book $B$, the sampling method is as follows: 
(1) Initialize a variable sliding window $w$ starting from $c_0$, expanding chapter by chapter.
(2) When $w=[c_j, c_{j+1}, \ldots, c_{k-1}, c_k]$ $(j \leq k < i)$:
   \begin{enumerate}[label=\roman*.,ref=\roman*,itemsep=0.001em, topsep=0.5em]
       \item \label{step1} If the length of $w < \inf R$, continue expanding $w$ with chapter $c_{k+1}$.
       \item \label{step2} If the length of $w \in R$, gather as a data point, set $w=[c_{k+1}]$.
       \item \label{step3} If the length of $w > \sup R$, remove chapters sequentially from $c_j$ onward until \ref{step1} or \ref{step2}.
   \end{enumerate}
(3) Follow the procedure until $B$ has no remaining chapters, then process the next book. This method is applied to each range $R$ across our book corpus, yielding four candidate sets.

Despite candidate sets being large, the sample volume gaps per book vary considerably. For example, books with longer total lengths but shorter chapters have more samples. Conversely, shorter books may have no or few samples in 3XL candidate set. We prioritize data from books with fewer samples to preserve the diversity. Regarding length, we prefer data closer to the target $T$. 
Table~\ref{bm_info} shows the final statistics of CNNSum. Each subset includes shorter data to maintain moderate length diversity, but the mean length is close to the target $T$, with the first quartile also kept near it. 
The final data volume is 695, based on two considerations: (1) The Annotation cost of long-context summaries. (2) Due to sample volume gaps across books, increasing the final volume will lead to imbalance, further introducing new biases from book content.

The key to long-context data is long dependencies within content, not just long sequence length. We assess using ProLong~\cite{longdependdata} and the results indicate that CNNSum performs strong long dependencies compared to other summarization datasets. Details are provided in Appendix~\ref{app:longdenpend_book}.

\subsection{Summary Annotation}
Our approach is similar to \textbf{Incremental Updating}~\cite{chang2023booookscore}, but with manual efforts. 
For a sample, we first generate a plot synopsis for each chapter using the prompt in Appendix~\ref{biaozhu_prompt}, mainly consisting of key plots and character information.
To reduce bias from a single model, we use various commercial LLMs as listed in Appendix~\ref{app:zhushiapis}.
Annotators then read each synopsis, select key plots by their judgment, and merge them into a final summary. If finding conflicts, annotators will ask LLMs to locate the source plots and correct them manually. 
We have two requirements for annotators: (1) Do not merely delete the unnecessary parts and merge the rest, but rewrite in your own words. (2) Avoid subjective commentary, focusing instead on objective plots. Although this may sacrifice some coherence, it increases the effective information density. For L and XL, we limit a maximum of 500 words, and 600 for 2XL and 3XL. Adding more words for coherence is not cost-effective, as most models do not produce overly long summaries. 

Our annotation team has 23 experts. To reduce individual bias, each sample in 2\textasciitilde3XL is annotated by one expert and then reviewed by another for consistency. This annotation strategy leverages the strengths of LLMs in efficiently processing multiple texts concurrently, alongside the logical thinking and error-correction abilities of human experts. This combination complements the weaknesses of each other, making it one of the best annotation strategies for long-context summarization tasks.

\begin{table*}[t]
  \centering
    \resizebox{0.98\linewidth}{!}{
  \begin{tabular}{l c cccc c}
    \hline
    \multirow{2}{*}{\textbf{Model}} & \multirow{2}{*}{\textbf{Ctx. Len.}} & \multicolumn{4}{c}{\textbf{CNNSum}} & \multirow{2}{*}{\textbf{MSE (\textit{P-IE vs P-IB})}}\\
     & & \textbf{L ($16k$)} & \textbf{XL ($32k$)} & \textbf{2XL ($64k$)} & \textbf{3XL ($128k$)} & \\
     
    \hline
    \multicolumn{7}{c}{\cellcolor{magenta!5}\textit{Commercial Models}} \\
    \hline
    GPT-4o-2024-08-06 & $128k$ & 15.5 & 14.2 & 12.5 & - & 0.0 \\
    GPT-4o-mini-2024-07-18 & $128k$ & 15.7 & 13.7 & 12.8 & - & 0.3 \\
    Gemini-1.5-pro & $2m$ & 19.3 & \uwave{18.1} & \underline{16.8} & \uwave{14.6} & 0.1 \\
    Qwen-plus-2024-09-19 & $128k$ & \underline{20.5} & \underline{18.5} & \uwave{16.4} & \underline{14.8} & 0.1 \\
    Doubao-pro-128k & $128k$ & \uwave{19.6} & 17.8 & 15.5 & 13.2 & 0.6 \\
    Moonshot-v1-128k & $128k$ & \textbf{22.4} & \textbf{20.3} & \textbf{18.0} & \textbf{15.2} & 0.9 \\
    
    \hline
    \multicolumn{7}{c}{\cellcolor{magenta!5}\textit{Open-Source Models}} \\
    \hline
    Yi-6B & $4k$ & 8.8 & 4.4 & 2.7 & 2.3 & 1.1 \\
    Yi-6B-Chat & $4k$ & 13.5 & 6.9 & 2.1 & 2.1 & 0.4 \\
    Yi-6B-200K & $200k$ & 9.9 & 9.4 & 8.8 & 4.0 & 4.5 \\
    Yi-34B-200K & $200k$ & 12.1 & 11.3 & 10.8 & 10.0 & 0.1 \\
    Yi-1.5-34B-32K & $32k$ & 11.6 & 10.5 & 9.6 & 0.1 & \textcolor{red}{14.5} \\
    Yi-1.5-34B-Chat-16K & $16k$ & 13.8 & 12.3 & 10.9 & 0 & \textcolor{red}{7.8} \\

    \hline
    InternLM2.5-7B-Chat-1M & $1m$ & 18.0 & 17.1 & 14.7 & 13.0 & 0.9 \\
    InternLM2.5-20B & $256k$ & 18.2 & 16.3 & 14.4 & 12.7 & 0.1 \\
    InternLM2.5-20B-Chat & $32k$ & \uwave{18.9} & \uwave{17.3} & \underline{16.2} & \underline{14.2} & 0.1 \\

    \hline
    ChatGLM3-6B-128K & $128k$ & 17.2 & 16.1 & \uwave{14.8} & \uwave{13.7} & 0.1 \\
    GLM4-9B-Chat-1M & $1m$ & \underline{19.0} & \textbf{17.9} & \textbf{16.5} & \textbf{15.4} & 0.2 \\

    \hline
    Llama3.1-8B & $128k$ & 7.8 & 8.0 & 8.4 & 3.1 & \textcolor{red}{7.2} \\
    Llama3.1-8B-Instruct & $128k$ & 15.6 & 14.3 & 12.8 & 9.9 & 1.4 \\
    Llama3.1-70B & $128k$ & 12.2 & 8.9 & 6.9 & 7.5 & \textcolor{red}{5.0} \\
    Llama3.1-70B-Instruct & $128k$ & 17.9 & 15.7 & 13.8 & 10.6 & 0.5 \\

    \hline
    LWM-Text-1M & $1m$ & 3.3 & 3.0 & 2.5 & 1.1 & 0.2 \\
    \hline
    Ministral-8B-Instruct-2410 & $128k$ & 16.2 & 14.0 & 11.3 & 3.5 & 0.1 \\

    \hline
    Qwen1.5-7B & $32k$ & 12.1 & 10.9 & 5.7 & 2.6 & \textcolor{red}{34.4} \\
    Qwen1.5-7B-Chat & $32k$ & 15.1 & 13.6 & 10.3 & 2.7 & 0.3 \\
    Qwen1.5-32B-Chat & $32k$ & 15.7 & 14.2 & 5.0 & 4.2 & 0.3 \\
    
    \hline
    Qwen2-7B & $128k$ & 9.8 & 8.4 & 8.0 & 6.9 & 1.5 \\
    Qwen2-7B-Instruct & $32k$ & 15.2 & 13.3 & 12.3 & 11.3 & 1.5 \\
    Qwen2-72B & $128k$ & 14.8 & 13.2 & 11.7 & 10.4 & \textcolor{red}{16.3} \\
    Qwen2-72B-Instruct & $32k$ & 15.5 & 13.2 & 11.5 & 9.2 & \textcolor{red}{19.0} \\

    \hline
    Qwen2.5-7B-Instruct & $32k$ & 15.7 & 14.0 & 12.7 & 9.6 & 3.4 \\
    Qwen2.5-72B-Instruct & $32k$ & \textbf{19.6} & \underline{17.6} & 13.6 & 13.4 & 1.3 \\
    
    \hline
  \end{tabular}
    }
    \vspace{-0.5em}
  \caption{Average ROUGE-L score of CNNSum. \textbf{Bold} denotes the best score, \underline{underline} for the second-best, and \uwave{wavy underline} for the third-best. \textcolor{red}{Red} highlights the large value ($\geq 5.0$) in \textbf{MSE (\textit{P-IE vs P-IB})}.}
  \label{bm_baseline_score_ave}
\vspace{-1em}
\end{table*}

\begin{table*}[t]
  \centering
    \resizebox{0.99\linewidth}{!}{
  \begin{tabular}{l c ccc | ccc | ccc | ccc}
    \hline
    \multirow{3}{*}{\textbf{Model}} & \multirow{3}{*}{\textbf{Ctx. Len.}} & \multicolumn{12}{c}{\textbf{CNNSum}}\\
     & & \multicolumn{3}{c}{\textbf{L ($16k$)}} & \multicolumn{3}{c}{\textbf{XL ($32k$)}} & \multicolumn{3}{c}{\textbf{2XL ($64k$)}} & \multicolumn{3}{c}{\textbf{3XL ($128k$)}}\\
     & & P & R & \textbf{F1} & P & R  & \textbf{F1} & P & R & \textbf{F1} & P & R & \textbf{F1}\\
    \hline
    GPT-4o-2024-08-06 & $128k$ & 61.2 & 59.4 & 60.3 & 60.9 & 59.3 & 60.1 & 60.1 & 56.8 & 58.0 & - & - & - \\
    GPT-4o-mini-2024-07-18 & $128k$ & 61.8 & 61.9 & 61.8 & 60.8 & 60.0 & 60.4 & 60.6 & 59.3 & 59.9 & - &- &-\\
    Gemini-1.5-pro & $2m$ & 66.6 & 63.9 & 65.1 & 66.6 & 64.4 & \underline{65.4} & 65.7 & 65.2 & \underline{65.4} & 65.0 & 65.3 & \textbf{65.1} \\
    Qwen-plus-2024-09-19 & $128k$ & 68.8 & 63.1 & \underline{65.7} & 67.9 & 61.6 & \uwave{64.5} & 66.6 & 60.8 & \uwave{63.5} & 65.3 & 60.3 & \uwave{62.6} \\
    Doubao-pro-128k & $128k$ & 67.3 & 63.1 & \uwave{65.0} & 66.1 & 60.6 & 63.1 & 64.7 & 58.0 & 61.1 & 61.5 & 55.6 & 58.3 \\
    Moonshot-v1-128k & $128k$ & 67.4 & 67.8 & \textbf{67.6} & 67.0 & 67.4 & \textbf{67.1} & 65.8 & 67.3 & \textbf{66.5} & 64.1 & 64.8 & \underline{64.4} \\
    
    \hline
  \end{tabular}
    }
    \vspace{-0.5em}
  \caption{Average BERTScore of commercial models. \textbf{Bold} denotes the best score, \underline{underline} for the second-best, and \uwave{wavy underline} for the third-best. \textbf{P} stands for precision, and \textbf{R} stands for recall.}
  \label{bm_baseline_btscore_ave}
\vspace{-0.5em}
\end{table*}

\section{Experiments}
\label{experiments}

\subsection{Baselines}
\label{sec:baselines_models}
Chinese has one of the largest corpora, but many advanced LLMs still support it poorly, as shown in Appendix~\ref{sec:llms_cibiao}.
Such as Claude3.5~\cite{anthropic2023claude} and GPT-4o cannot process 3XL. We select models as follows; details are in Appendix~\ref{sec:bm_models_details}.

\paragraph{Commercial Models} 
(1) OpenAI flagship, GPT-4o-2024-08-06~\cite{openai2024} and the lite version GPT-4o-mini-2024-07-18~\cite{openai2024mini}. (2) Moonshot-v1-128k~\cite{moonshot2024}. (3) Qwen-plus-2024-09-19~\cite{aliyun2024}. (4) Doubao-pro-128k~\cite{volcengine2024}. All the above have a 128k context length. (5) Gemini-1.5-pro~\cite{team2024gemini}, with 2 million context length.

\paragraph{Open-Source Models} 
\label{open_source_models}
(1) Yi~\cite{young2024yi} and Yi-1.5 series. (2) Qwen1.5, Qwen2~\cite{yang2024qwen2}, and Qwen2.5~\cite{qwen2.5} series. (3) ChatGLM3-6B-128K~\cite{du2021glm} and special version GLM4-9B-Chat-1M~\cite{glm2024chatglm}. (4) InternLM2.5-20B~\cite{cai2024internlm2} series and special version InternLM2.5-7B-Chat-1M. (5) Llama3.1 series~\cite{dubey2024llama}. (6) Ministral-8B-Instruct-2410~\cite{mistral2023ministraux}. (7) LWM-Text-1M~\cite{liu2024world}.

\subsection{Experimental Setup}
We set generation tokens to 400 for L and XL, and 500 for 2\textasciitilde3XL. We use jieba\footnote{\url{https://github.com/fxsjy/jieba}} for Chinese segmentation and then report ROUGE-L~\cite{lin2004rouge}.
For the BERTScore~\cite{zhangbertscore}, we use the Chinese XLNet-Large~\cite{cui-etal-2020-revisiting}. This architecture can handle long texts, and the computation is based on the hidden states of the last layer.

\paragraph{Baseline Evaluation}
We use vLLM~\cite{kwon2023efficient} for open-source models and apply greedy sampling.  We set $temperature=0$ for commercial APIs.
The details can be seen in Appendix~\ref{sec:baseline_setup}. The prompts we used are shown in Appendix~\ref{two_baseline_prompts}.

\paragraph{Fine-tuning Data}
Our dataset contains about 9,000 novel summarization samples, independent of the CNNSum corpus. The length ranges from 2k\textasciitilde4k with Yi tokenizer. The prompts are shown in Appendix~\ref{sec:two_train_prompts}. We randomly concatenated the samples, similar to Section~\ref{sec:bm_length_settings}, setting ranges to 14k\textasciitilde18k and 30k\textasciitilde34k, with average lengths of about 16k and 32k. The main motivation is to enable models to adapt to more positions and activate extrapolation ability.~\cite{tian2024untie} contemporaneously proposes a similar data augmentation strategy, proving it enhances long-context ability.

\paragraph{Fine-tuning Experiments} 
We follow findings from~\cite{chen2023longlora}, use LoRA (rank=8)~\cite{hu2021lora}, and unfreeze embedding and normalization layers. Each experiment is repeated at least three times. More details are in Appendix~\ref{sec:training_setup}.

\begin{table}[t]
  \centering
  \setlength{\tabcolsep}{0.4em}  
\begin{adjustbox}{max width=\linewidth}
\begin{tabular}{l c cccc}
    \hline
    \multirow{2}{*}{\textbf{Model}} & \multirow{2}{*}{\textbf{Ctx. Len.}} & \multicolumn{4}{c}{\textbf{CNNSum}} \\
     & & \textbf{L} & \textbf{XL} & \textbf{2XL} & \textbf{3XL}\\
    \hline
    & & \graycell{\textbf{14.4}} & \graycell{\textbf{12.7}} & \graycell{\textbf{10.8}} & \graycell{0.1}\\
    \multirow{-2}{*}{Yi-1.5-34B-32K} & \multirow{-2}{*}{$32k$} & 8.8 & 8.2 & 8.3 & 0.1 \\
    & & \graycell{\textbf{15.4}} & \graycell{\textbf{14.2}} & \graycell{\textbf{12.1}} & \graycell{0.0}\\
    \multirow{-2}{*}{Yi-1.5-34B-Chat-16K} & \multirow{-2}{*}{$16k$} & 12.1 & 10.4 & 9.7 & 0.0 \\

    \hline
    & & \graycell{6.1} & \graycell{6.6} & \graycell{7.0} & \graycell{2.2}\\
    \multirow{-2}{*}{Llama3.1-8B} & \multirow{-2}{*}{$128k$} & \textbf{9.4} & \textbf{9.4} & \textbf{9.7} & \textbf{3.9} \\
    & & \graycell{11.8} & \graycell{7.5} & \graycell{5.2} & \graycell{7.0}\\
    \multirow{-2}{*}{Llama3.1-70B} & \multirow{-2}{*}{$128k$} & \textbf{12.5} & \textbf{10.2} & \textbf{8.6} & \textbf{7.9} \\

    \hline
    & & \graycell{\textbf{16.5}} & \graycell{\textbf{14.5}} & \graycell{\textbf{7.0}} & \graycell{2.5}\\
    \multirow{-2}{*}{Qwen1.5-7B} & \multirow{-2}{*}{$32k$} & 7.7 & 7.2 & 4.4 & \textbf{2.6} \\
    
    \hline
    & & \graycell{12.1} & \graycell{10.7} & \graycell{10.3} & \graycell{9.5}\\
    \multirow{-2}{*}{Qwen2-72B} & \multirow{-2}{*}{$128k$} & \textbf{17.5} & \textbf{15.7} & \textbf{13.1} & \textbf{11.3} \\
    & & \graycell{12.4} & \graycell{10.5} & \graycell{10.6} & \graycell{\textbf{10.6}}\\
    \multirow{-2}{*}{Qwen2-72B-Instruct} & \multirow{-2}{*}{$32k$} & \textbf{18.5} & \textbf{15.8} & \textbf{12.3} & 7.8 \\
    
    \hline
    \end{tabular}
    \end{adjustbox}
      \vspace{-0.5em}
  \caption{\colorbox{lightgray!50}{Gray} for Prompt-IE and white for Prompt-IB. \textbf{Bold} denotes the best score of each model.}  \label{bm_baseline_two_prompt_some}
\vspace{-0.5em}
\end{table}

\subsection{Main Results on CNNSum}
\label{sec:main_results}
We calculate mean ROUGE and use Mean Squared Error (MSE) to measure performance gaps caused by prompts, as shown in Table~\ref{bm_baseline_score_ave} (Full results are shown in Appendix~\ref{sec:bm_baseline_score_all}). Overall, commercial models outperform most open-source models. The low MSE demonstrates a more stable long-context summarization ability. Moonshot-v1 performs the best, but it suffers a decrease of 32.1\% at 3XL. Gemini-pro, Qwen-plus, and Doubao-pro have similar performance. Doubao-pro shows the largest decrease of 32.7\% from L to 3XL, while Gemini-pro shows the smallest decrease of 24.4\%. Unexpectedly, GPT-4o performs the worst, lagging behind other models by 20\%\textasciitilde30\% on average from L to 2XL. 
Given that the output quality of commercial APIs is better, and their ROUGE scores are close, we specifically compute the BERTScore for them, as shown in Table~\ref{bm_baseline_btscore_ave}.  However, they almost have the same trend, providing limited additional meaningful insights.  This indicates the limitations of simple automatic metrics.  Therefore, in the subsequent parts, we only use ROUGE as the basic guiding metric and conduct detailed human inspection.

GLM4-9B-Chat-1M performs best in the open-source models, with only an 18.9\% decrease from L to 3XL. InternLM2.5-20B-Chat shows the most robust extrapolation ability to 3XL, with a 24.8\% decrease. In the Qwen series, Qwen2.5-72B-Instruct performs the best but decreases by 31.6\% when extrapolated to 3XL.
Others perform similarly on L and XL, with the Qwen2 series decreasing 20\%\textasciitilde40\% to 3XL. Due to lower Chinese encoding efficiency, Llama3.1 and Ministral require extrapolation on 3XL, with Llama3.1 series showing an average decrease exceeding 40\% and Ministral reaching as high as 78.3\%. Yi-34B-200K generally outputs just half or one sentence or meaningless repetitions. And Yi-6B-200K performs even worse.

\subsection{Analysis and Case Study}
\label{bm_analysis}
\paragraph{Why Do GPT-4o Fail?}
As shown in Table~\ref{bm_info} and Table~\ref{bm_baseline_btscore_ave}, GPT-4o actually needs to process longer sequences. Despite this, its performance on L still trails other commercial models on XL on average by 16\%. GPT-4o and Qwen-plus generate more concise summaries, while Moonshot-v1 and Gemini-pro generate longer ones. Moonshot-v1 outputs significantly longer than others, with many samples reaching the length limit without repetition. Longer outputs contain richer information, are more likely to overlap with references, leading to higher scores. For GPT-4o, its already shorter outputs often include subjective commentary. These contents are extensions of the plot rather than itself, usually not overlapping with references, as shown in Appendix~\ref{gpt4o_fail}. 
On the other hand, such highly generalized content may indicate that the model has limited memory of key plots.
Other commercial models rarely exhibit this and focus on objective plots, regardless of the output length.


\paragraph{Which Prompt? Which Model Version?}
\label{bm_prompt_analysis}
Results of models with large MSE (in red) are detailed in Table~\ref{bm_baseline_two_prompt_some}.
Models with shorter context lengths perform better with Prompt-IE, namely Qwen1.5 and Yi-1.5.
We inspect their outputs and find that the instruction may be forgotten after reading long contexts in Prompt-IB. Many outputs copy from novels or are meaningless repetitions. Conversely, using Prompt-IE can summarize normally even in extrapolation, examples are in Appendix~\ref{prompt_vs_ie_vs_ib}.
The Base model is more stable than its Chat version with Prompt-IE, which is not directly reflected by ROUGE. Yi-1.5-34B generates more concise and clean summaries, while the Chat model outputs more content, but it is more prone to disorganization and repetition. This observation also occurs in models with lower MSE, such as InternLM2.5-20B series. Examples are shown in Appendix~\ref{prompt_vs_base_vs_chat}.

Models with longer context lengths seem to recognize the instruction at the beginning easily. However, it still may forget the instruction when extrapolating on a very long context with Prompt-IB, such as Qwen2-72B-Instruction. Examples are in Appendix~\ref{prompt_vs_qwen272b}. Llama3.1 is more sensitive to prompt types. They sometimes even fail to produce any meaningful content with Prompt-IE, see Appendix~\ref{prompt_vs_llama31}; this is more frequent in longer samples. But when using Prompt-IB, it does not occur at all. 

The types of abnormal outputs are summarized below:
(1) Initially outputs normally, but at random moments it starts repeating a sentence or paragraph.
(2) After generating a summary, it fails to stop and starts repeating extra prompt instructions.
(3) Copy from novels randomly until reaching the length limit.
(4) Fail to follow the instructions and output randomly. This may depend on the prompt template.
(5) Crash directly and repeat meaninglessly.
(6) A combination of the above situations.
The repetition issues partly arise from a greedy sampling strategy~\cite{meister2023typicalsmaple, xu2022learningrepe}, but they actually reflect the ability deficiencies in handling long text. 
On the one hand, these stem from the training phase configuration, including data volume of summarization tasks, prompt diversity, and so on.
On the other hand, the primary reason is the challenge of length extrapolation. When position embeddings exceed the maximum training sequence length, it may lead to a breakdown in the attention mechanism~\cite{han2024lminfinite}. Although most baseline LLMs have been extended for extrapolation during pretraining, their stability in summarization tasks remains inadequate.

\paragraph{Rise of Small Long-Context LLMs}
In the InternLM2.5 series, 7B-1M slightly outperforms 20B and achieves 94\% of 20B-Chat average performance. Similar observations are in Qwen, GPT-4o, and Llama3.1.
Despite larger LLMs having better reasoning ability, they hardly apply to long-context novel plots. 
In Table~\ref{bm_baseline_btscore_ave}, Gemini-pro, which supports longer context, maintains a high recall across all subsets, indicating its stronger ability to memorize the key plots. In contrast, GPT-4o shows a significant decline.
We summarize the primary challenges: (1) A longer context length means stable memory ability for key plots. (2) Models with supervised fine-tuning better recognize and follow the instructions after reading long novel texts.
For many samples, we observe that GPT-4o summarizes with identical structures and plots to GPT-4o mini, differing only in phrasing, as shown in Appendix~\ref{large_model_vs_small_model_gpt4o}.
Training small long-context models is significantly less challenging and cheaper, and thus they are cost-effective for novel summarization.

\begin{figure}[t]
  \setlength{\tabcolsep}{0.4em}  
    \centering
    \begin{adjustbox}{max width=\linewidth}
  \begin{tabular}{l ccc cccc}
    \hline
    \multirow{2}{*}{\textbf{Base Model}} & \multicolumn{3}{c}{\textbf{CLongEval-LStSum}} & \multicolumn{4}{c}{\textbf{CNNSum}}\\
     & \textbf{Small} &\textbf{Medium} &\textbf{Large} & \textbf{L} & \textbf{XL} & \textbf{2XL} & \textbf{3XL}\\
    \hline
     & \graycell{17.2}  & \graycell{\textbf{15.2}} & \graycell{10.0} & \graycell{\textbf{18.4}} & \graycell{\textbf{16.1}} & \graycell{11.4} & \graycell{\textbf{3.1}}\\
     \multirow{-2}{*}{Qwen1.5-7B} & \textbf{17.4} & 14.8 & \textbf{11.1} & 18.1 & 15.7 & \textbf{12.1} & 2.9 \\
     & \graycell{16.7}  & \graycell{14.7} & \graycell{7.4} & \graycell{17.7} & \graycell{15.6} & \graycell{8.5} & \graycell{2.6}\\
     \multirow{-2}{*}{Qwen1.5-7B-Chat} & 16.8 & 14.2 & 8.8 & 17.2 & 15.0 & 9.2 & 2.6\\
     \hline
     & \graycell{\textbf{17.2}}  & \graycell{14.8} & \graycell{11.9} & \graycell{16.8} & \graycell{\textbf{15.6}} & \graycell{12.4} & \graycell{\textbf{9.0}}\\
     \multirow{-2}{*}{Yi-6B} & 17.0 & \textbf{15.0} & 12.0 & \textbf{17.1} & 15.5 & \textbf{12.5} & 8.3\\
     & \graycell{15.5}  & \graycell{14.6} & \graycell{\textbf{12.9}} & \graycell{16.6} & \graycell{15.5} & \graycell{\textbf{12.5}} & \graycell{6.0}\\
     \multirow{-2}{*}{Yi-6B-Chat} & 15.6 & 14.5 & 12.4 & 16.1 & 15.5 & 12.3 & 6.0\\
    \hline
  \end{tabular}
  \end{adjustbox}
    \vspace{-0.5em}
  \captionof{table}{\colorbox{lightgray!50}{Gray} for Prompt-IE, white for Prompt-IB. \textbf{Bold} denotes the best score within each model series.}
  \label{tab:train_two_prompt_basechat_model}
      \vspace{0.5em}
    \includegraphics[width=0.99\linewidth]{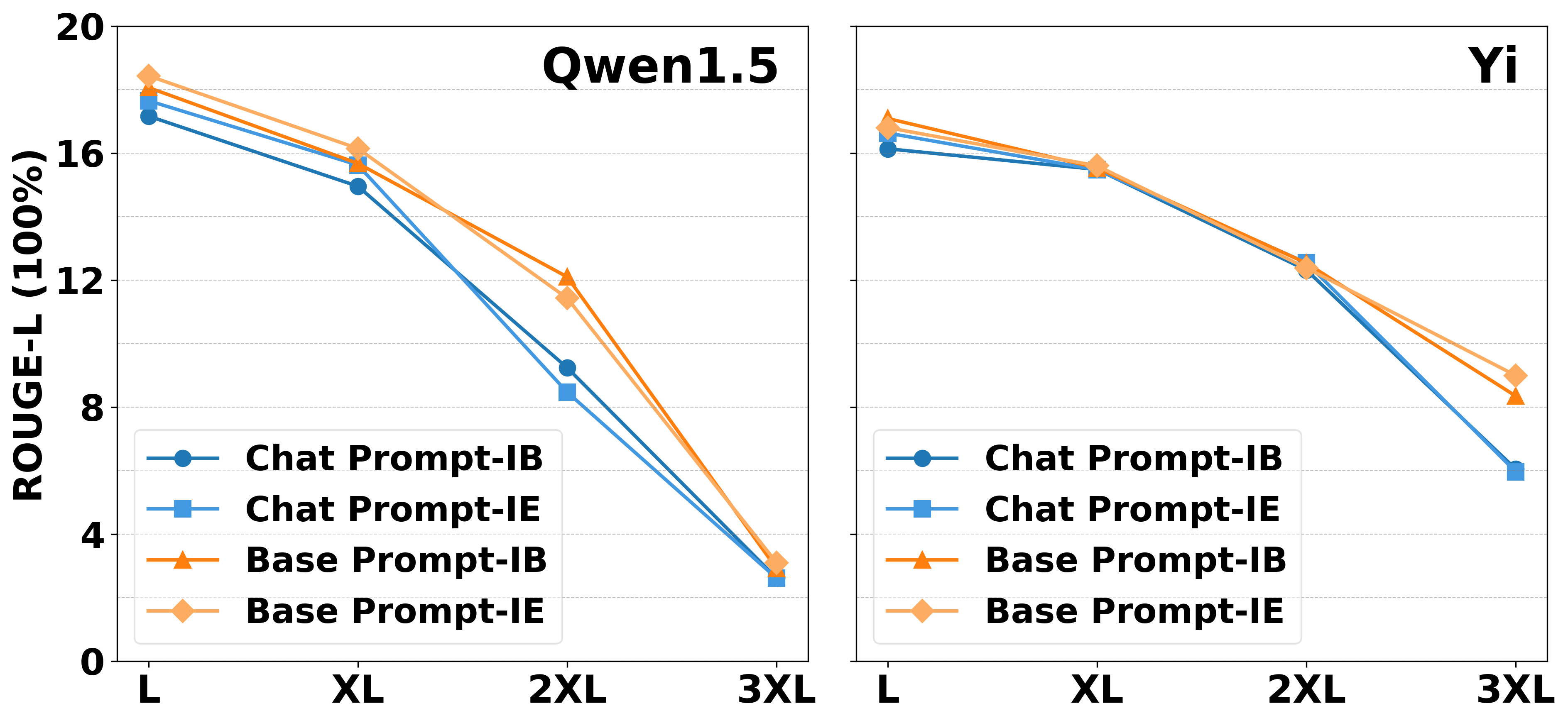}
    \vspace{-2em}
    \captionof{figure}{Results of fine-tuned models on CNNSum.}
    \label{fig:train_two_prompt_basechat_model}
    \vspace{-1em}
\end{figure}



\subsection{Ablation Study for Fine-Tuning}

We have performed analyses based on various general LLMs. However, fine-tuning powerful models for summarization tasks needs further exploration. We fine-tune Yi-6B series and Qwen1.5-7B series on our training sets with an average length of 16k and evaluate on CNNSum and CLongEval-LStSum~\cite{qiu2024clongeval}. CLongEval has three subsets: Small (1k\textasciitilde16k), Medium (16k\textasciitilde50k), and Large (50k\textasciitilde100k). Unlike CNNSum, its sample length distribution is uniform. More statistics are in Appendix~\ref{app:clongeval_length}.
Results are shown in Table~\ref{tab:train_two_prompt_basechat_model}.

\paragraph{Extrapolation Potential}
Compared to Table~\ref{bm_baseline_score_ave}, the performance of Yi-6B on L and XL significantly improves by 2\textasciitilde3 times, while Qwen1.5 also improves by 50\%.
Despite Yi-6B having the shortest default context length (4k), its RoPE base is scaled up via ABF~\cite{xiong2023effective}, enabling greater potential for long context.
In contrast, the improvement in Qwen1.5 mainly comes from the enhanced ability to follow instructions for summarization tasks, reducing the meaningless outputs.
With the largest RoPE base in our baselines (details in Appendix~\ref{sec:bm_models_details}), Yi-6B shows better extrapolation potential on 2\textasciitilde3XL.


\begin{figure*}[t]
    \centering
    \begin{minipage}[b]{0.496\textwidth}
        \centering
        \includegraphics[width=0.52\textwidth]{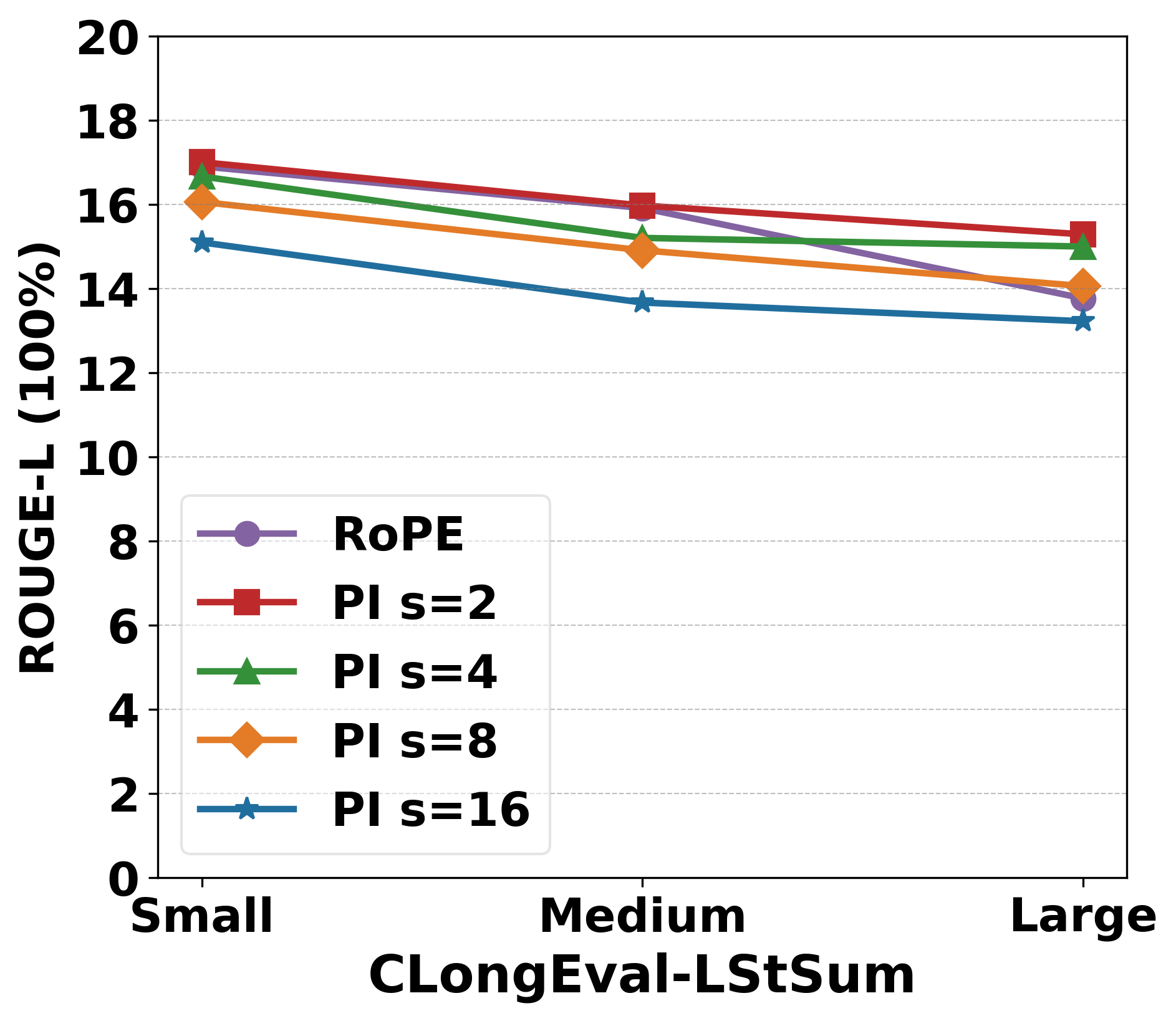}
        \hspace{-0.035\textwidth} 
        \includegraphics[width=0.485\textwidth]{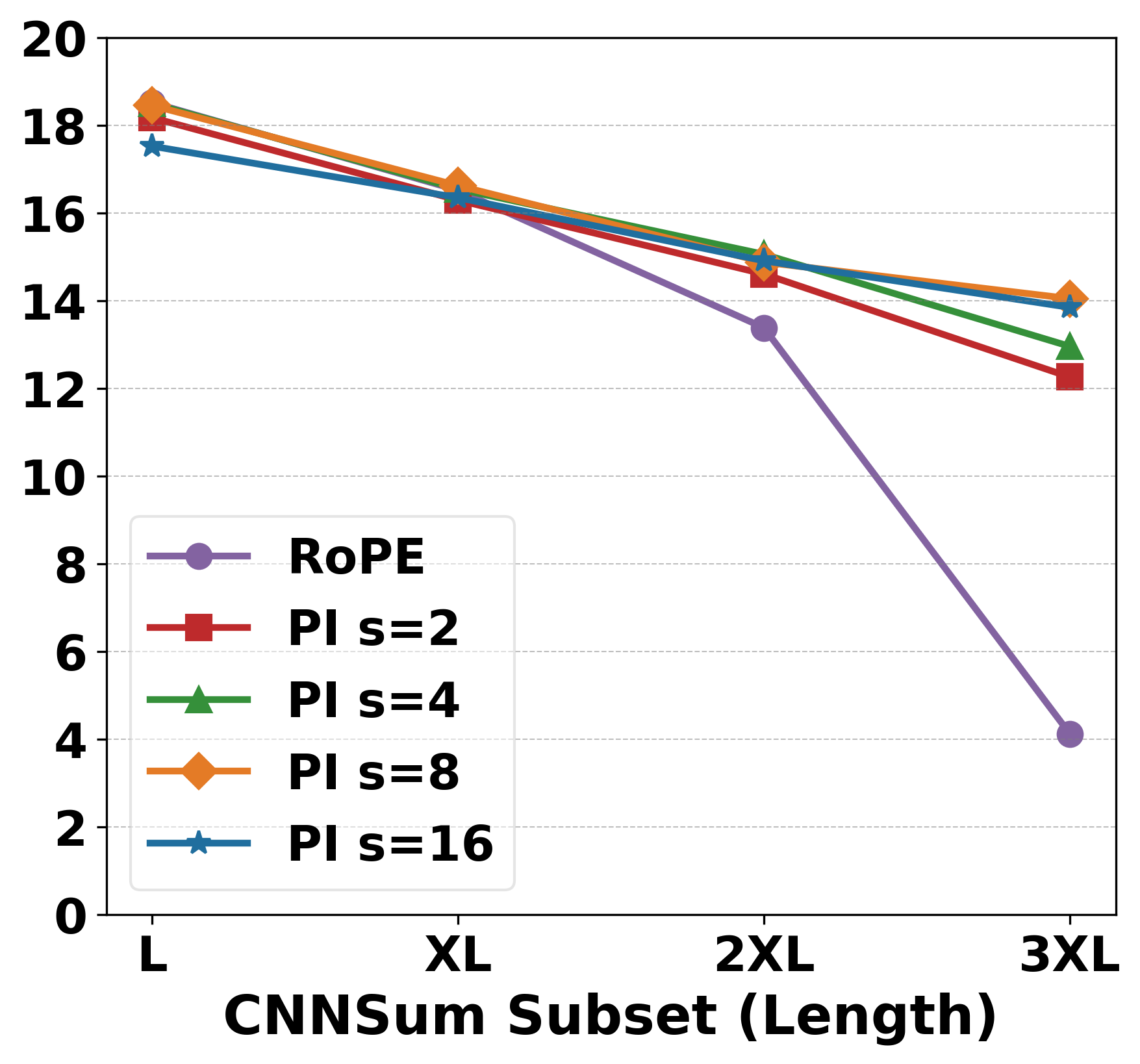}
        \vspace{-2em}
        \caption*{(a) Train Average Len. $32k$ with PI}
    \end{minipage}
    \hfill
    \begin{minipage}[b]{0.496\textwidth}
        \centering
        \includegraphics[width=0.52\textwidth]{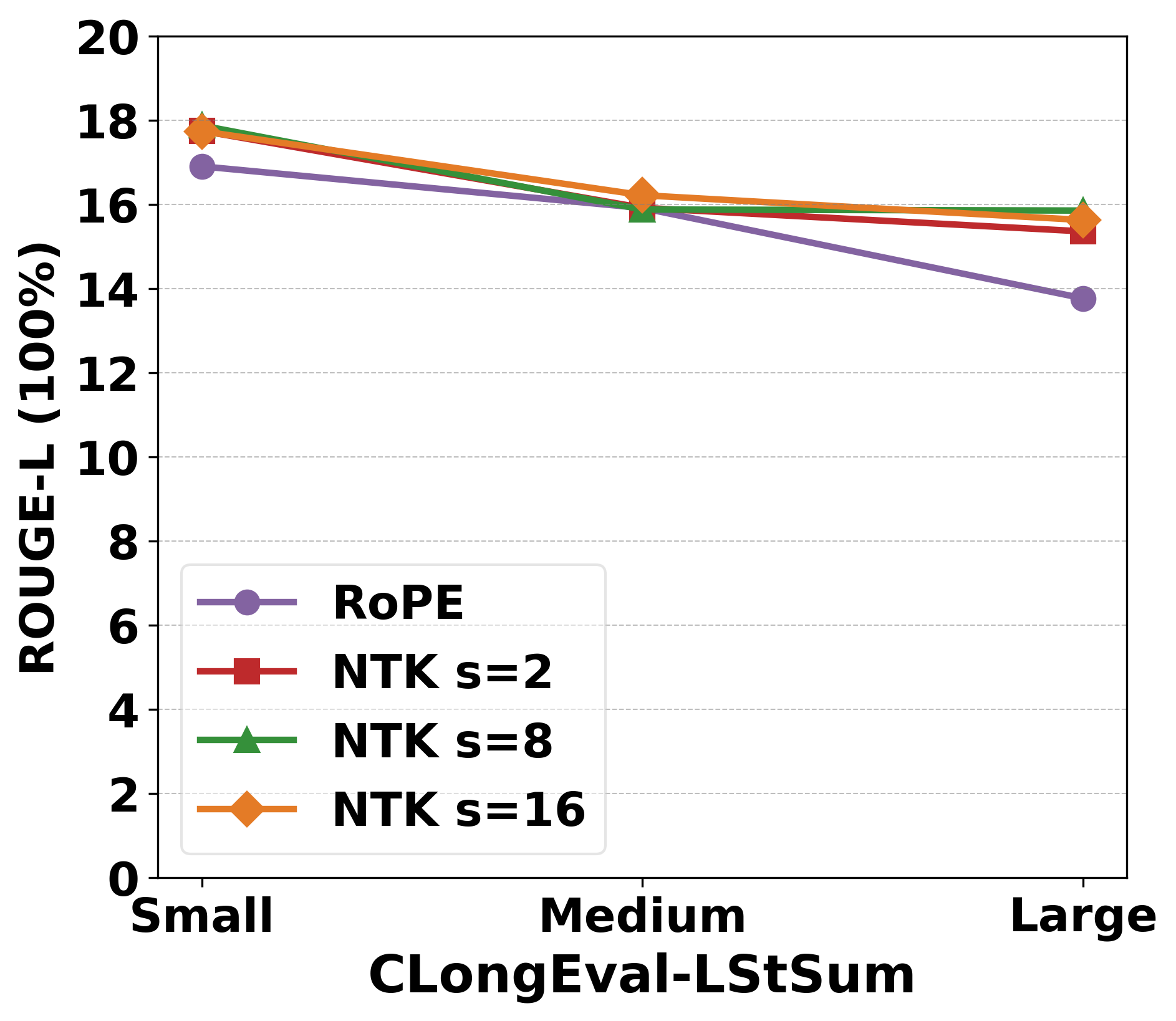}
        \hspace{-0.035\textwidth} 
        \includegraphics[width=0.485\textwidth]{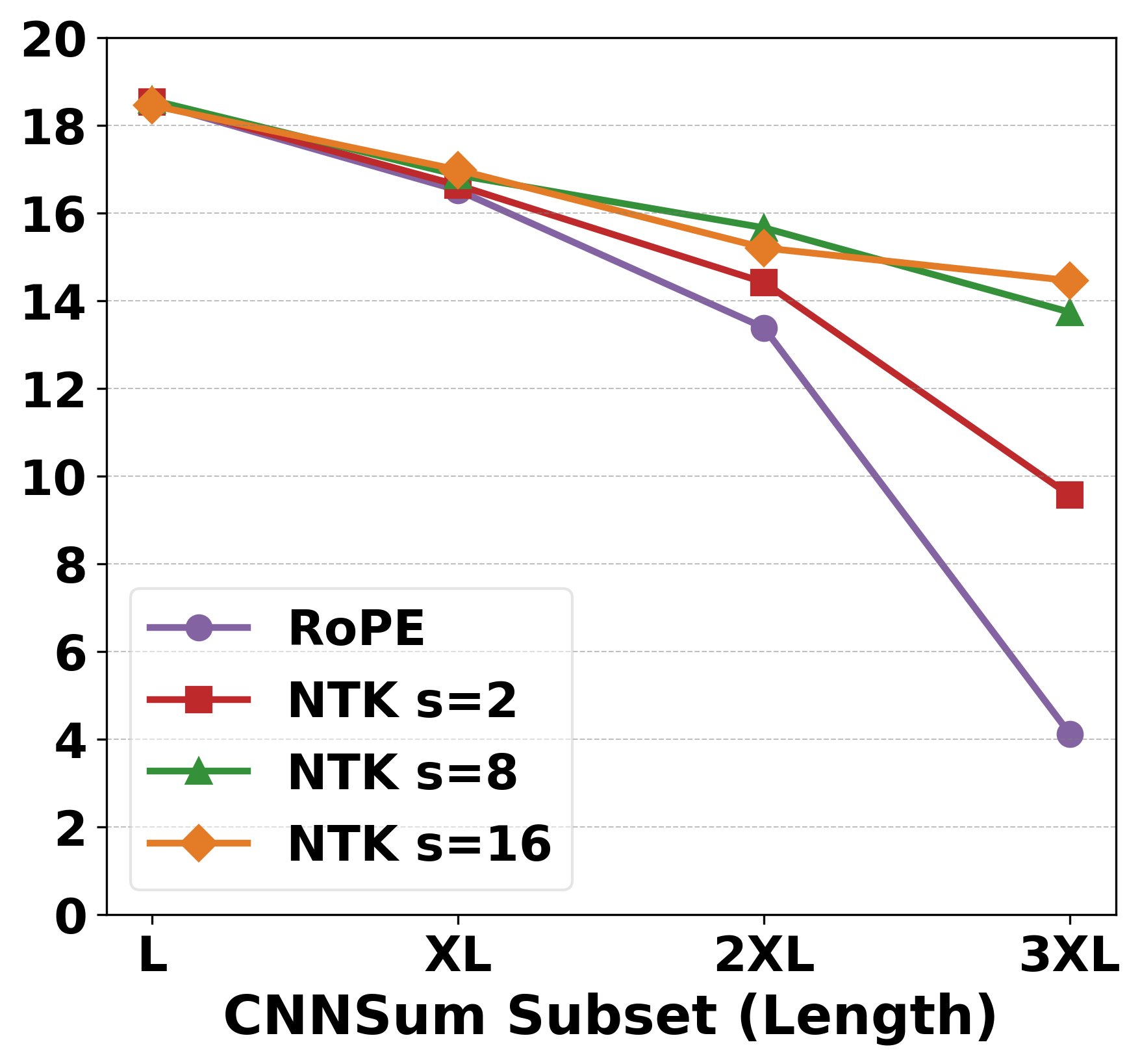}
        \vspace{-2em}
        \caption*{(b) Train Average Len. $32k$ with NTK}
    \end{minipage}
    \vspace{-2em}
    \caption{Results comparison of Fine-tuned Qwen1.5-7B on CNNSum and CLongEval-LStSum.}
    \label{fig:qwen_train_rope}
    \vspace{-1em}
\end{figure*}


\paragraph{Model Version}
Within each model series, the best scores for most subsets are achieved by the Base version. Fig.~\ref{fig:train_two_prompt_basechat_model} indicates that the Base models demonstrate better extrapolation potential and are more suitable for efficient fine-tuning.
For example, we observe that Qwen1.5-7B-Chat exhibits worse stability when extrapolating on 2XL, having more completely meaningless repetition.

\paragraph{Prompt-IB vs Prompt-IE}
In Table~\ref{bm_baseline_two_prompt_some}, Qwen1.5-7B shows the largest performance gaps. However, these gaps nearly disappear after fine-tuning.
This is confirmed by Fig.~\ref{fig:train_two_prompt_basechat_model}, which intuitively shows that a fine-tuned model has consistent performance, regardless of the prompt templates. We check that the outputs have no significant difference in quality.


\begin{figure}[t]
    \centering
    \includegraphics[width=0.518\linewidth]{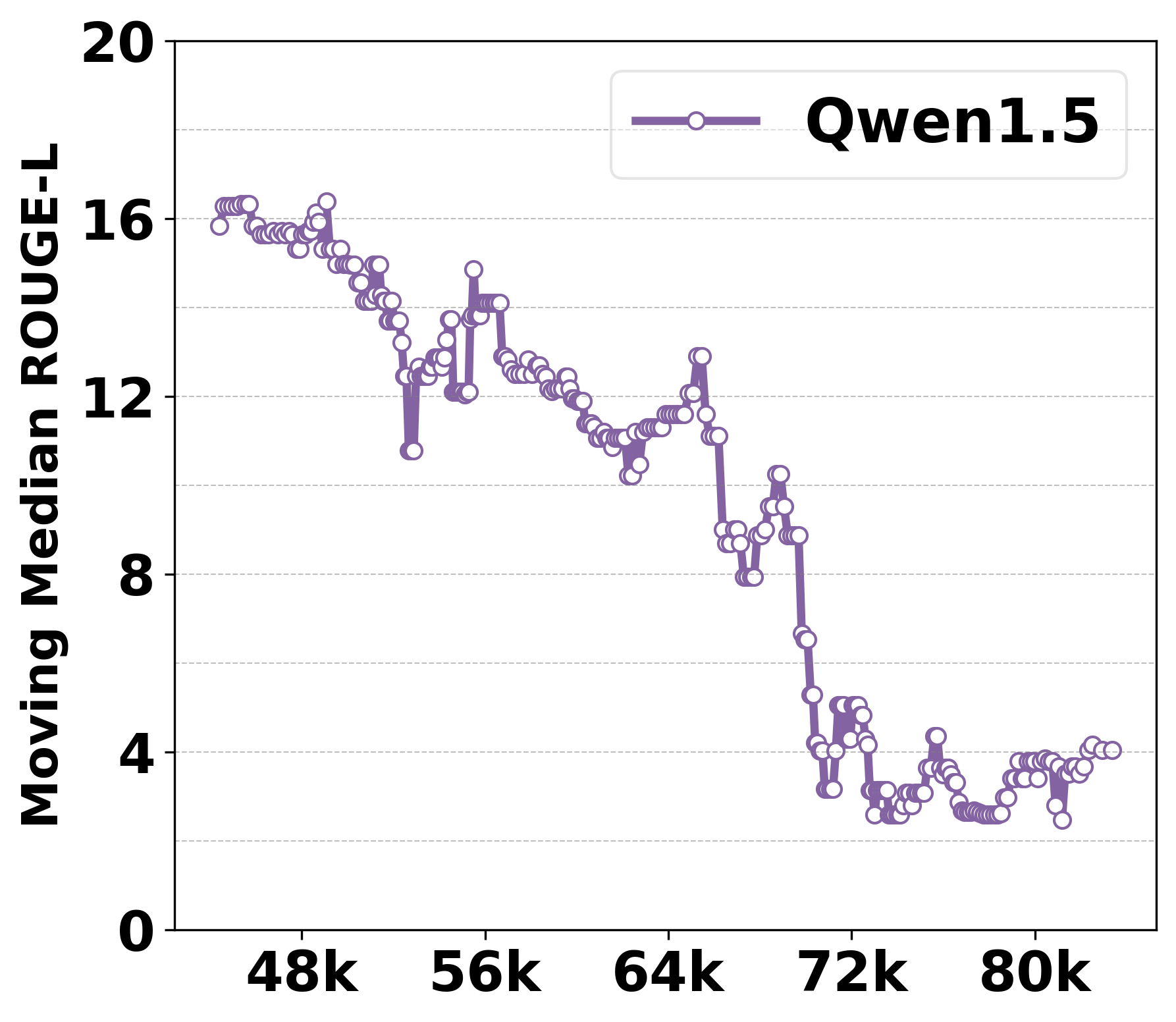}
    \hspace{-0.014\textwidth} 
    \includegraphics[width=0.486\linewidth]{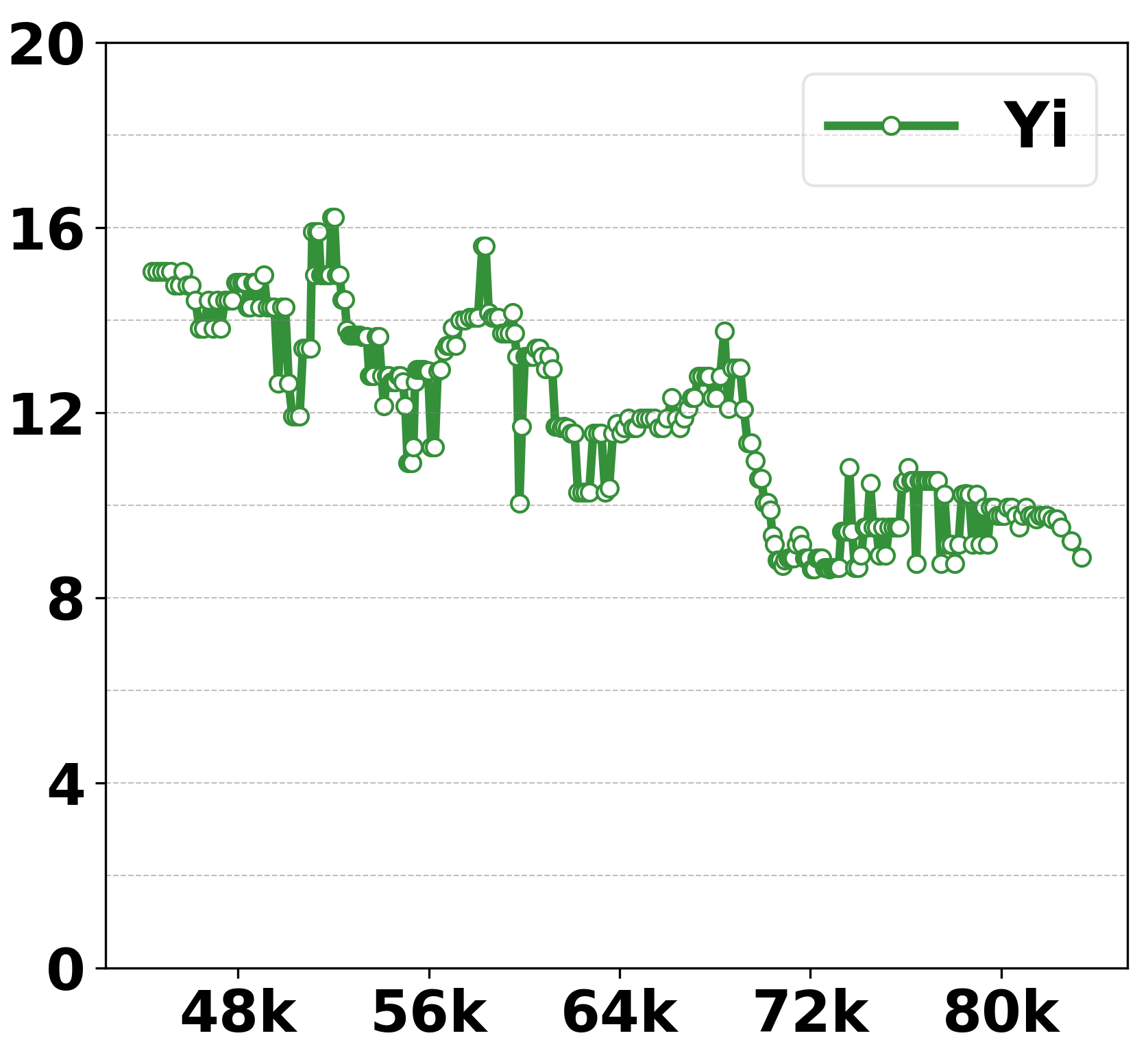}
    \vspace{-2em}
    \caption{Moving Median ROUGE-L (window size=15) of fine-tuned models on CLongEval-LStSum Large}
    \label{fig:qwen15_clongeval}
    \vspace{-1em}
\end{figure}

\subsection{Reliable Evaluation of Extrapolation}
\label{reliable_evaluation}
In Table~\ref{tab:train_two_prompt_basechat_model}, the scores of CLongEval-LStSum Large (50k\textasciitilde100k) and 2XL (64k) are higher than 3XL (128k).
We find the output quality has decreased significantly on 2XL, with the type (1) summarized in Section~\ref{bm_prompt_analysis} occurring frequently.
The shorter samples have much better output, leading to misleading final scores, as shown in Fig.~\ref{fig:qwen15_clongeval}. To further evaluate long-context summarization extrapolation reliably, we fine-tune models using RoPE scaling methods. Full results can be seen in Appendix~\ref{sec:train_baserope_result}. 

The misleading nature of CLongEval further increases. In Fig.~\ref{fig:qwen_train_rope} (a), CLongEval shows that original RoPE and PI~\cite{chen2023pi} perform comparably, obtaining high scores. 
PI demonstrates a slight decrease in the overall score as $s$ increases, which is a known limitation. 
Conversely, CNNSum demonstrates that the extrapolation ability of the original RoPE is weaker, leading to a 66\% decrease on 3XL. 
PI with larger $s$ performs poorer on L and XL, but as context increases, smaller $s$ shows a faster decrease due to insufficient interpolation.
In Fig.~\ref{fig:qwen_train_rope} (b), CNNSum also clearly shows more significant differences. 
Although PI theoretically cannot extrapolate on length without training due to distance resolution variations, models with ABF seem to adapt. NTK~\cite{reddit_ntkaware_scaled_rope} interpolates more moderately than PI, leading to a faster decrease with the same $s$ on 3XL (e.g., $s=2$). 

\begin{table}[t]
\vspace{0.6em}
  \setlength{\tabcolsep}{0.25em}  
  \centering
    \begin{adjustbox}{max width=1\linewidth}
  \begin{tabular}{l ccc cccc}
    \hline
    \multirow{2}{*}{\textbf{Train Len. (Ave.)}} & \multicolumn{2}{c}{\textbf{YaRN$(s=16)$}} & \multicolumn{4}{c}{\textbf{CNNSum}}\\
     & \textbf{$\alpha(Slow)$} & \textbf{$\beta(Fast)$} & \textbf{L} & \textbf{XL} & \textbf{2XL} & \textbf{3XL}\\
    \hline
    \multirow{6}{*}{$14k\sim18k$ ($16k$)} & \graycell{1} & \graycell{32} & 17.5 & \textbf{16.0} & \textbf{14.1} & 5.8 \\
       & 1 & 4 & \textbf{17.7} & \textbf{16.0} & 13.6 & 5.7 \\
       & 0.1 & 32 & 17.0 & 14.3 & 11.5 & 5.5\\
       \cline{2-7}
       & \multicolumn{2}{c}{\textbf{RoPE}} & 16.8 & 15.6 & 12.4 & \textbf{9.0} \\
        \cline{2-7}
       & \multicolumn{2}{c}{\textbf{PI ($s=16$)}} & 16.1 & 13.9 & 12.0 & 4.1 \\
        \cline{2-7}
       & \multicolumn{2}{c}{\textbf{NTK ($s=16$)}} & 16.2 & 14.7 & 11.3 & 5.1 \\
    \hline
  \end{tabular}
  \end{adjustbox}
    \vspace{-0.5em}
  \caption{\colorbox{lightgray!50}{Gray} denotes YaRN default settings. Low $\beta$ means more no-interpolation dimensions, low $\alpha$ means less "low-frequency" dimensions using PI.}
  \label{train_yarn_16k}
\vspace{-1em}
\end{table}

\begin{figure*}[t]
    \centering
    \begin{minipage}[b]{0.2531\textwidth}
        \centering
        \includegraphics[width=\textwidth]{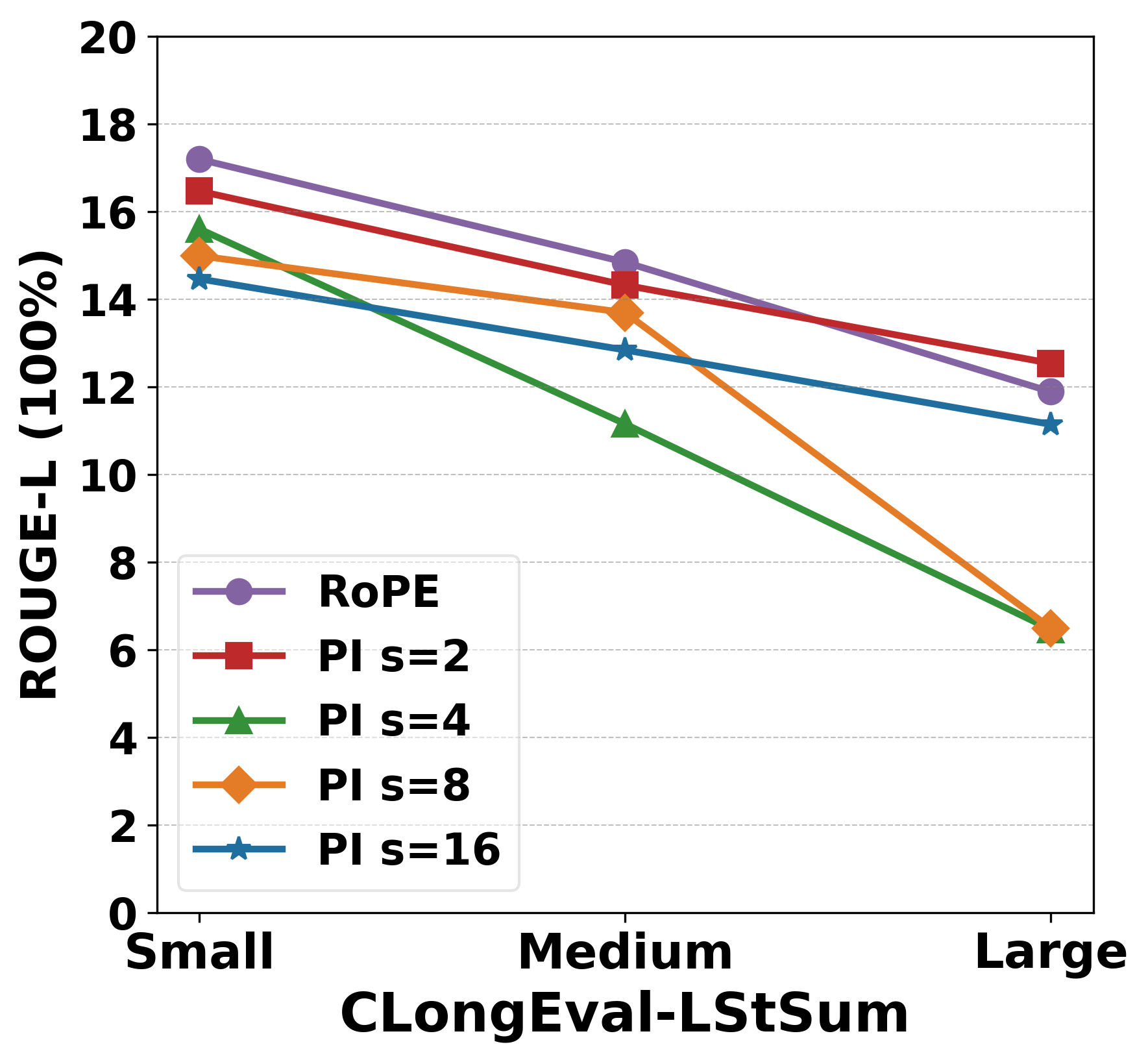}
    \end{minipage}%
    \hspace{-0.015\textwidth} 
    \begin{minipage}[b]{0.2531\textwidth}
        \centering
        \includegraphics[width=\textwidth]{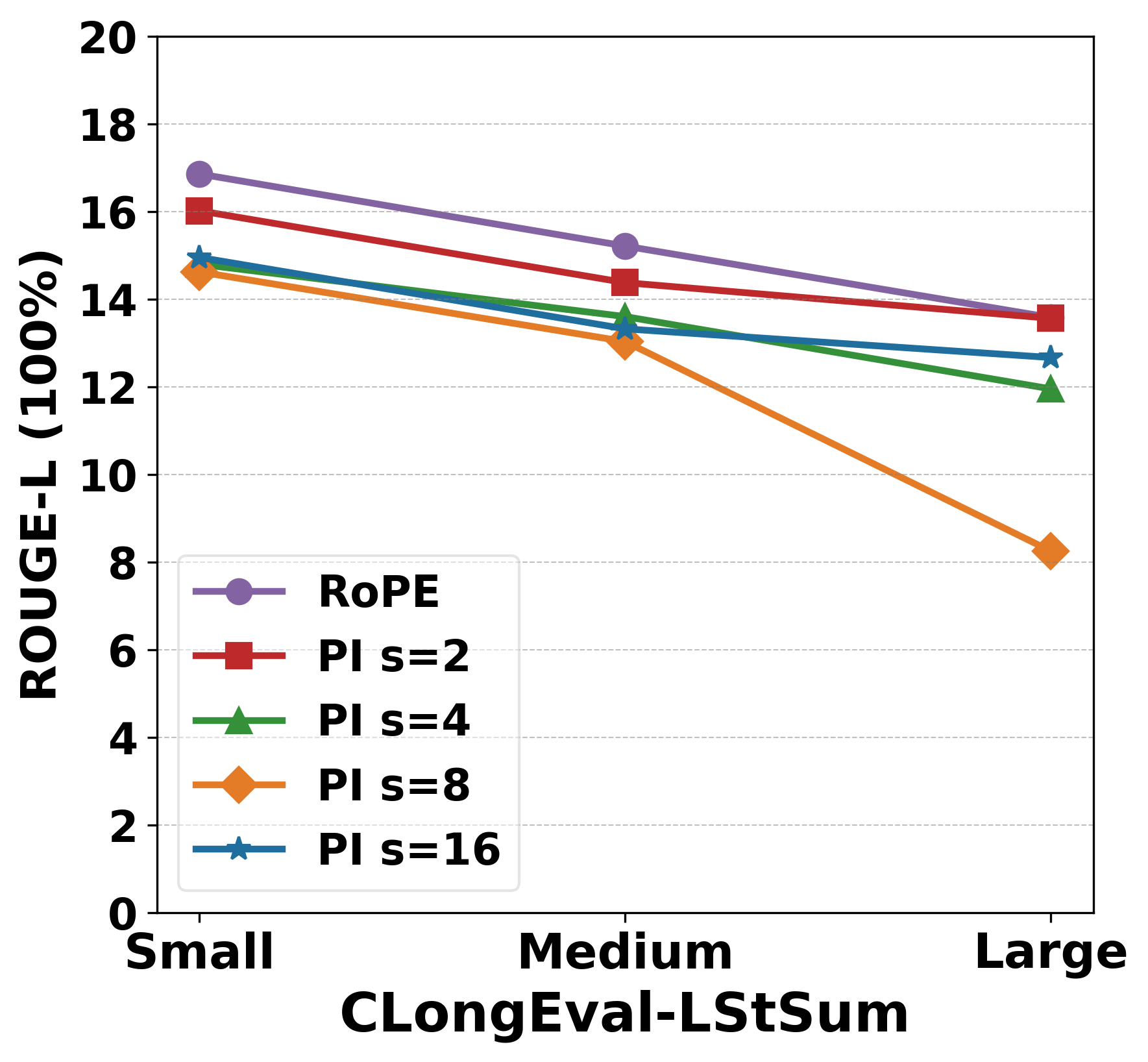}
    \end{minipage}
    \begin{minipage}[b]{0.2531\textwidth}
        \centering
        \includegraphics[width=\textwidth]{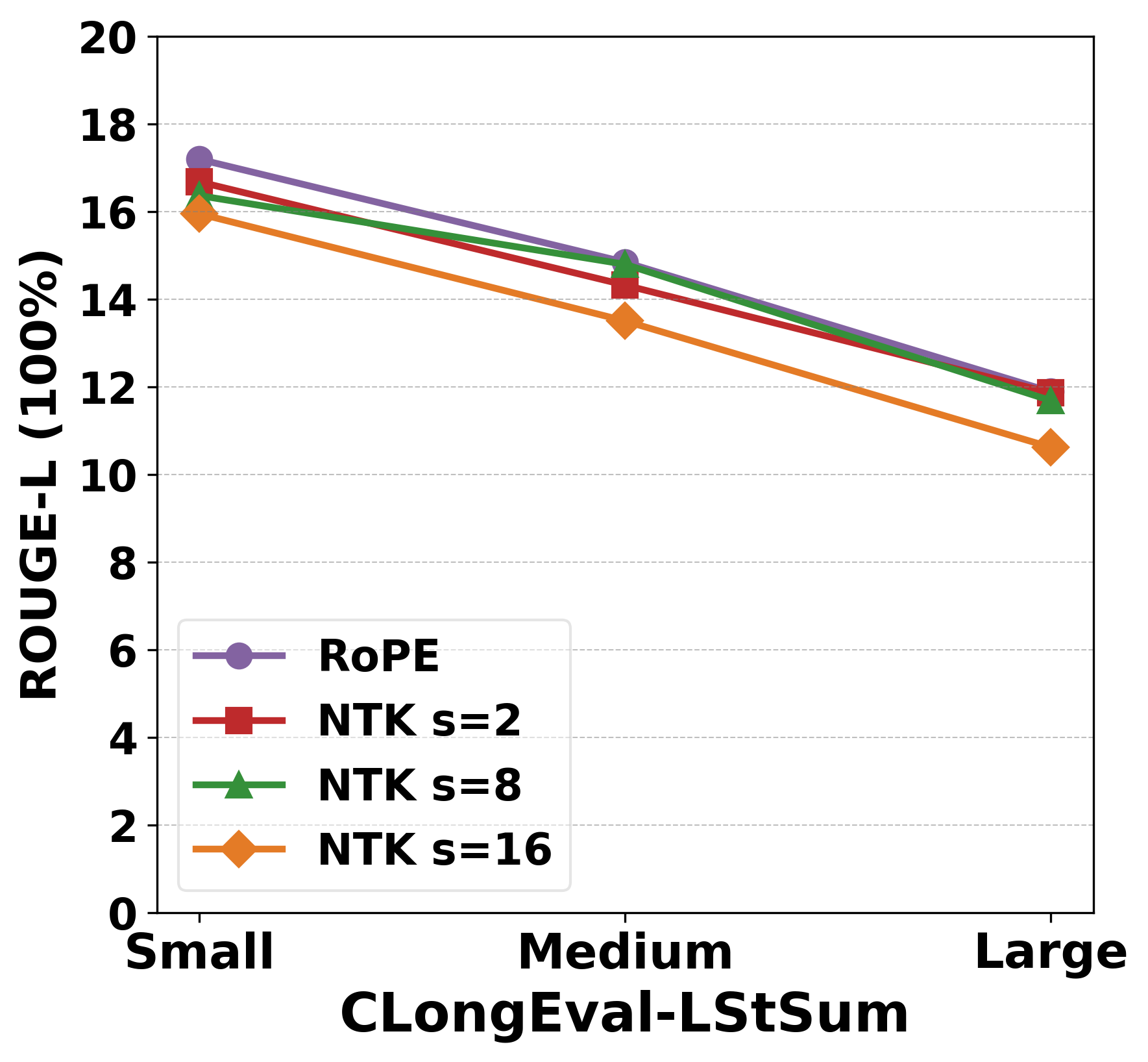}
    \end{minipage}
    \hspace{-0.02\textwidth} 
    \begin{minipage}[b]{0.2531\textwidth}
        \centering
        \includegraphics[width=\textwidth]{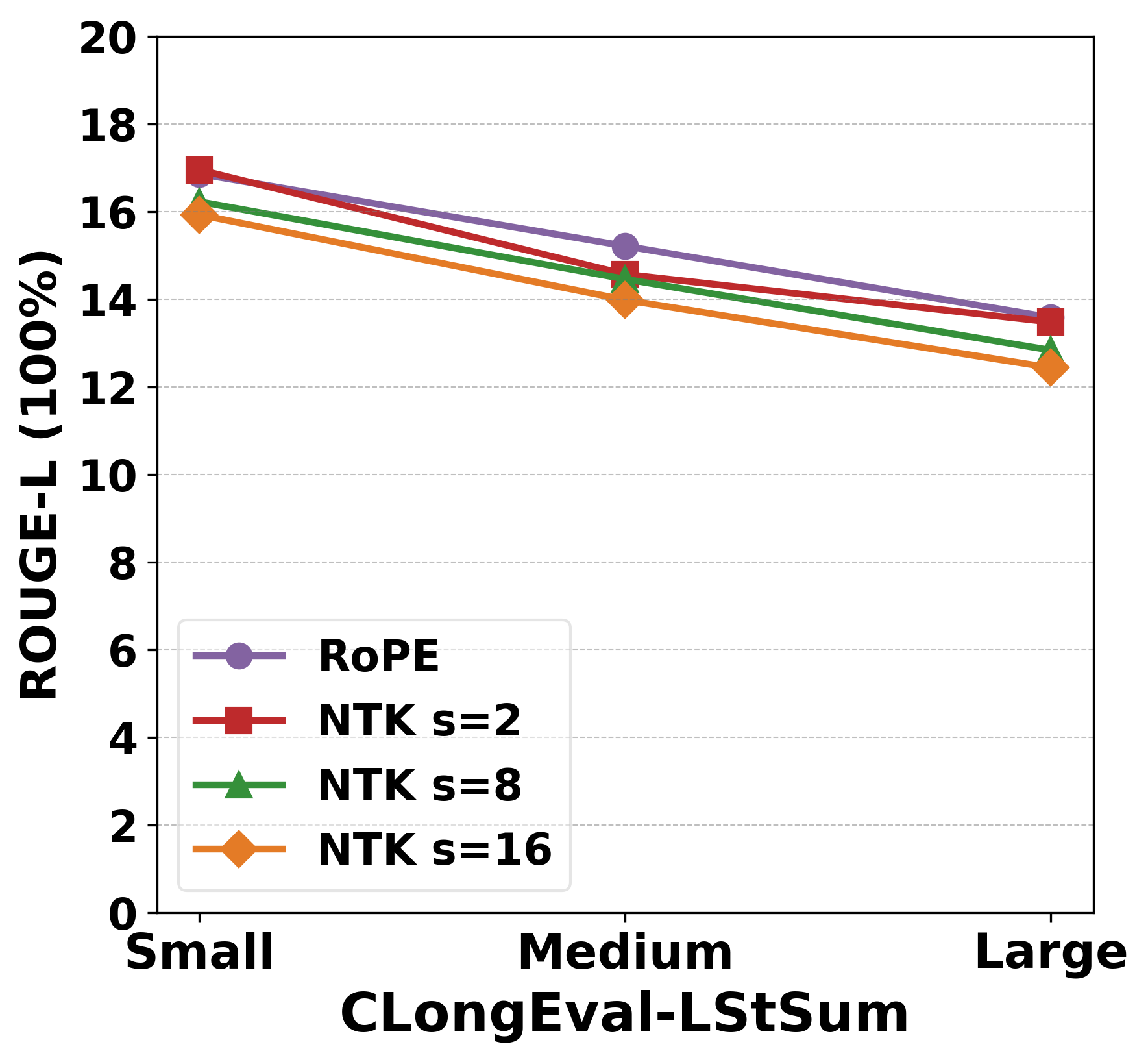}
    \end{minipage}
    

    \begin{minipage}[b]{0.249\textwidth}
        \centering
        \includegraphics[width=\textwidth]{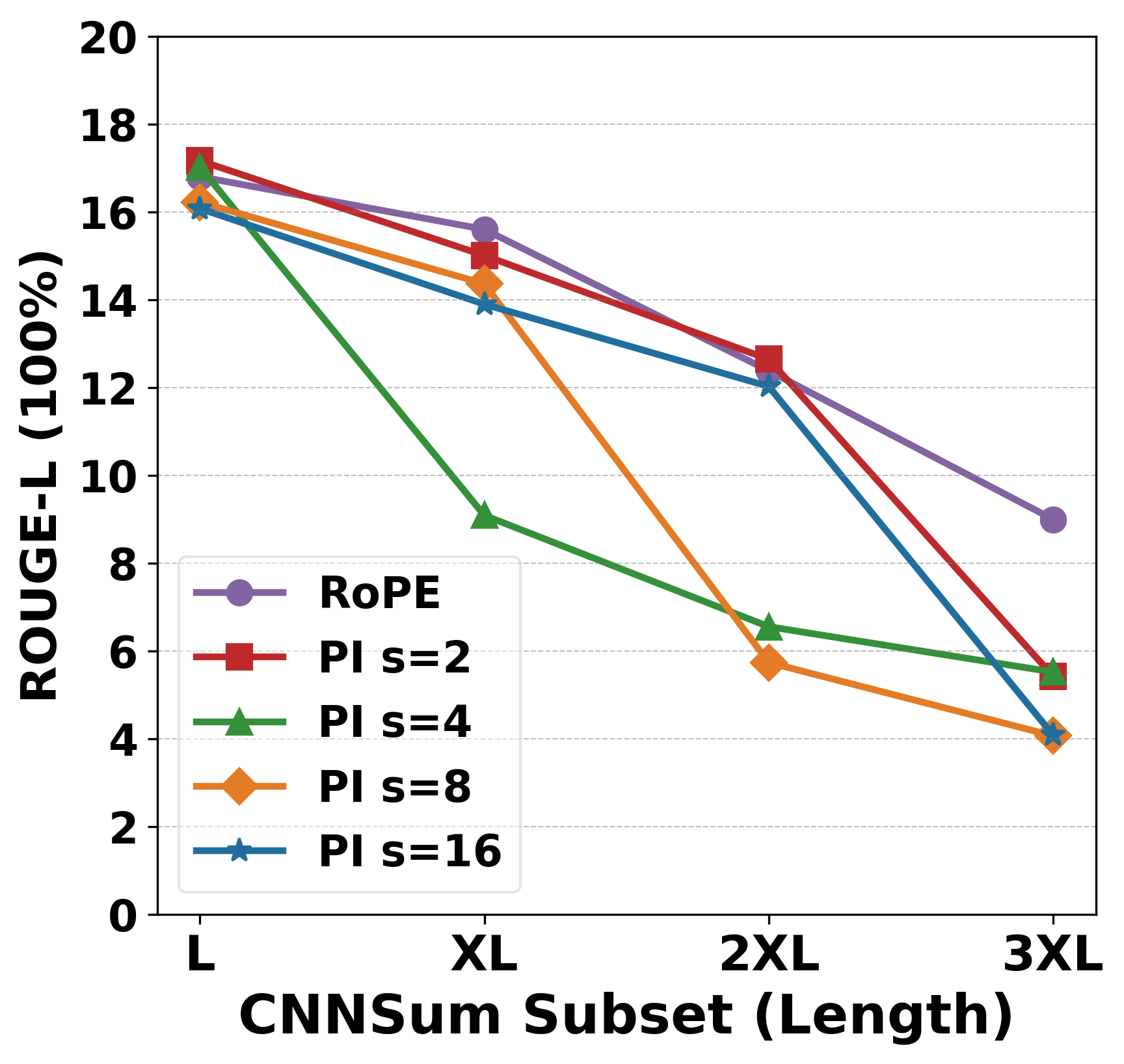}        
        \vspace{-1.5em}
        \caption*{(a) Train Ave. Len. $16k$}
    \end{minipage}%
    \hspace{-0.01\textwidth} 
    \begin{minipage}[b]{0.249\textwidth}
        \centering
        \includegraphics[width=\textwidth]{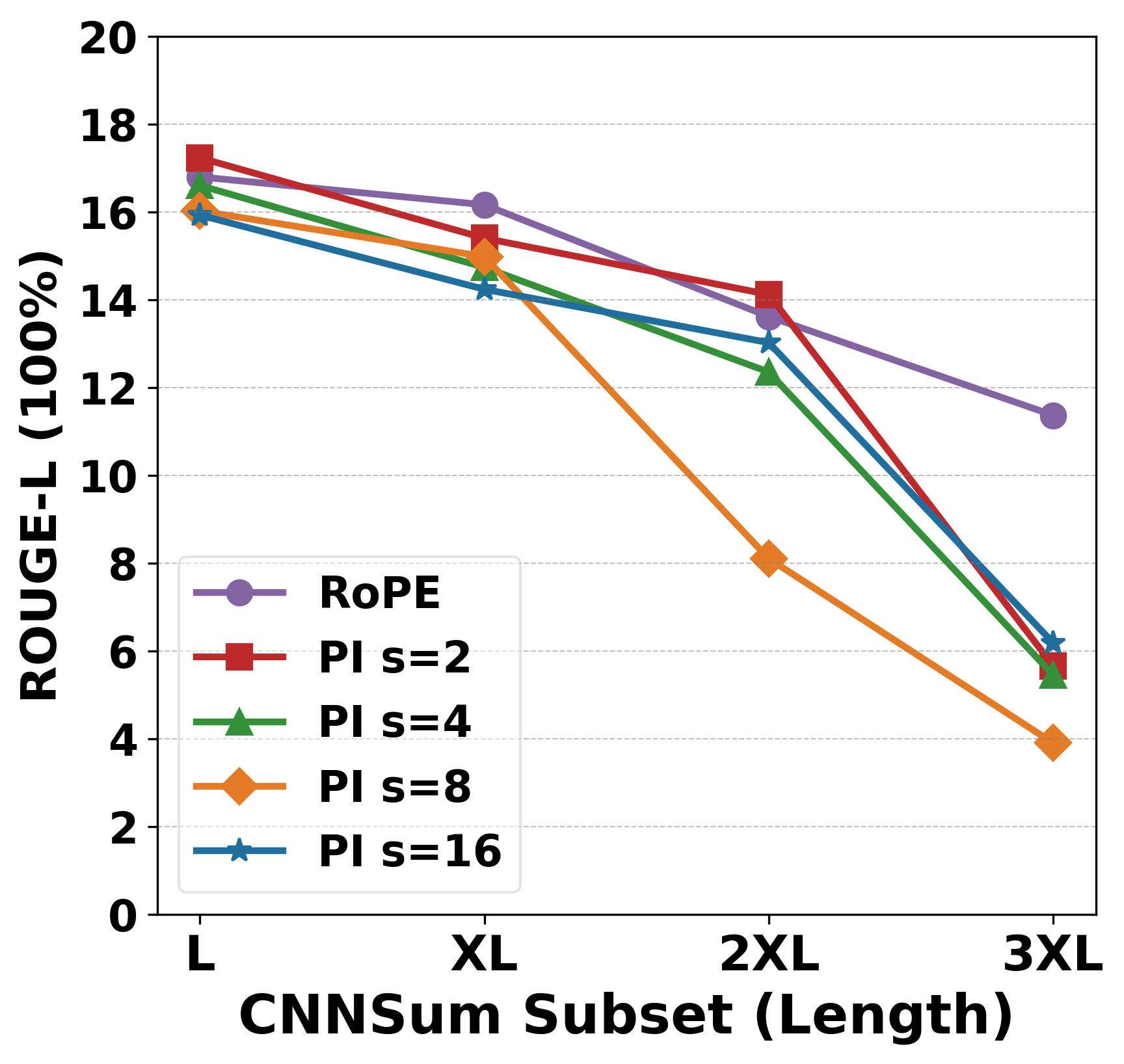}
        \vspace{-1.5em}
        \caption*{(b) Train Ave. Len. $32k$}
    \end{minipage}
    \begin{minipage}[b]{0.249\textwidth}
        \centering
        \includegraphics[width=\textwidth]{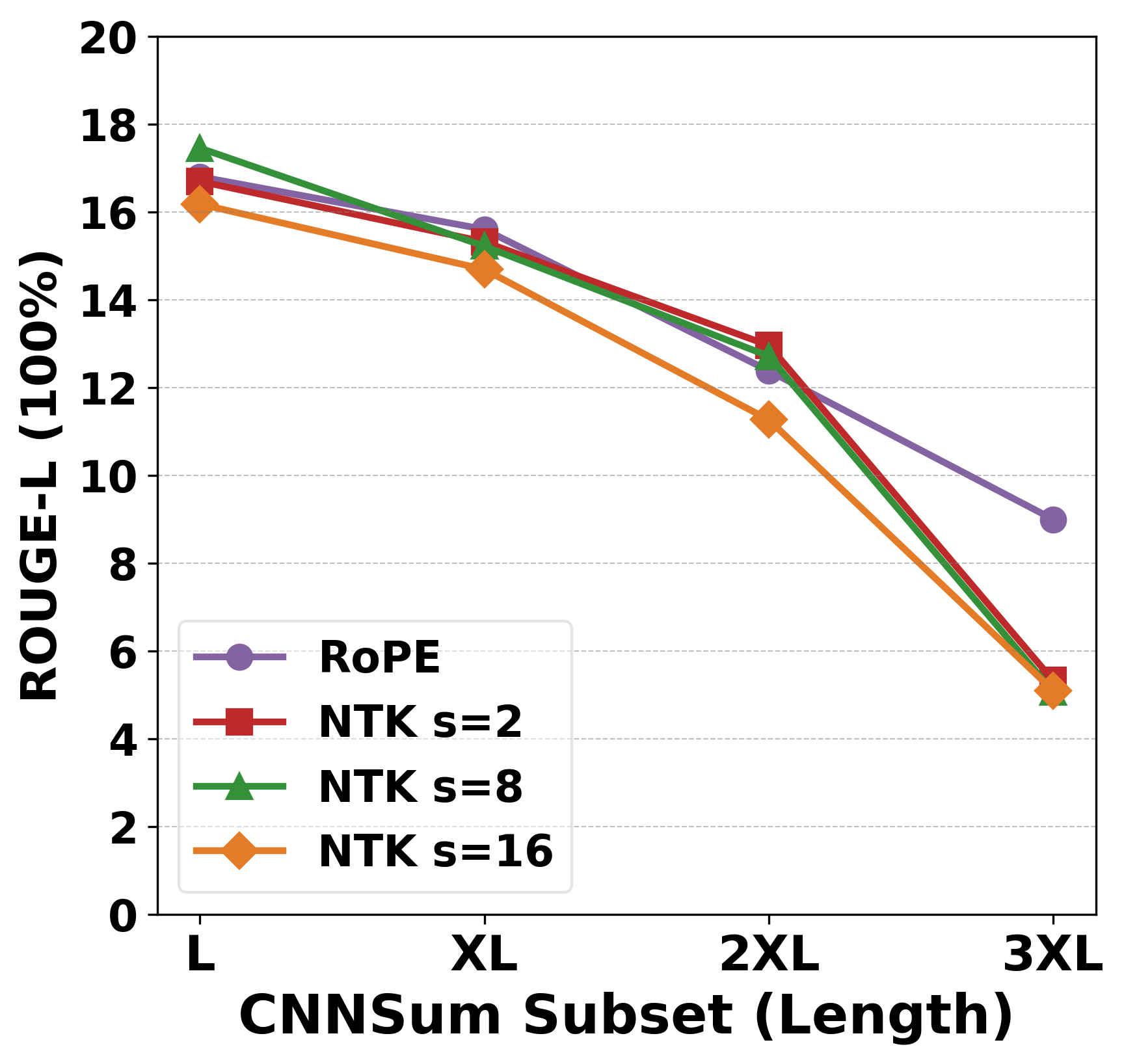}
        \vspace{-1.5em}
        \caption*{(c) Train Ave. Len. $16k$}
    \end{minipage}
    \hspace{-0.014\textwidth} 
    \begin{minipage}[b]{0.249\textwidth}
        \centering
        \includegraphics[width=\textwidth]{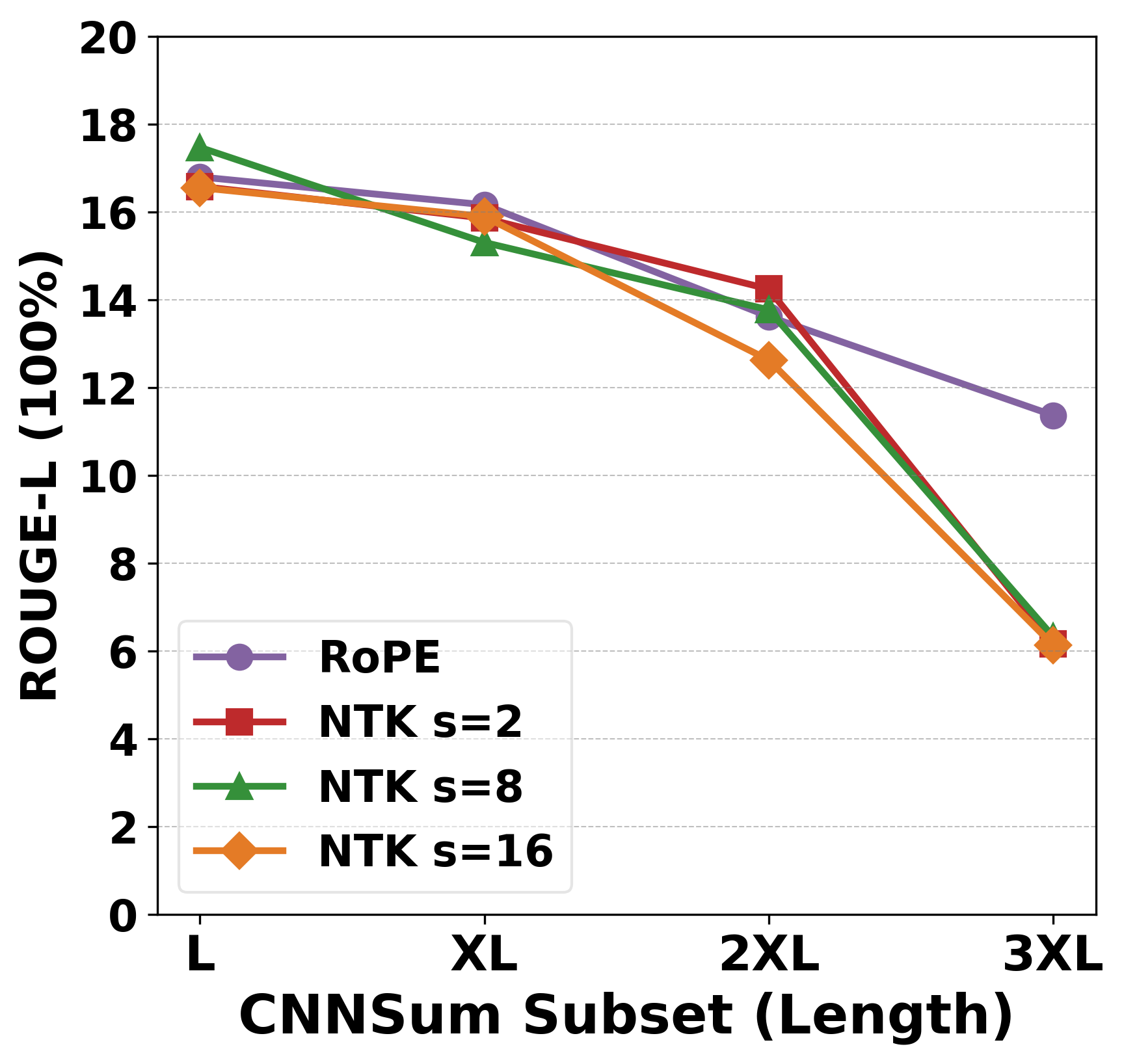}
        \vspace{-1.5em}
        \caption*{(d) Train Ave. Len. $32k$}
    \end{minipage}
    \vspace{-0.5em} 
    \caption{Results comparison of Fine-tuned Yi-6B on CNNSum and CLongEval-LStSum.}
    \label{fig:yi_train_rope}
    \vspace{-1em}
\end{figure*}

Yi-6B shows some interesting differences. The original RoPE with a larger base performs best on 3XL. In Fig.~\ref{fig:yi_train_rope} (a) and (b), CLongEval shows that PI exhibits an abnormal decrease. CNNSum provides a clear explanation. In Fig.~\ref{fig:yi_train_rope} (a), PI with $s=2$ extrapolates to 2XL (64k). But for $s=4, 8, 16$, extrapolate respectively to 16k, 32k, and 64k, exactly matching interpolation based on its default 4k context length.
Fig.~\ref{fig:yi_train_rope} (c) and (d) indicate it is more sensitive to high-frequency information of RoPE in extrapolation.
The larger base (50,000,000) significantly diminishes the rotation speed of RoPE.
With the default settings of YaRN~\cite{peng2023yarn}, 58\% "low-frequency" dimensions (37/64) use PI, and only 20\% "high-frequency" dimensions (13/64) are without interpolation.
Reducing interpolation leads to slight improvements, as shown in Table~\ref{train_yarn_16k}. 

These observations suggest that interpolation methods should be used cautiously for models with RoPE-base scaled.
Fine-tuning without interpolation appears to be stable and straightforward for models with a large RoPE.
We also have training observations that differ from~\cite{peng2023yarn}. With the same scale $s$, PI led to a higher initial loss and slower convergence than NTK and original RoPE. This grows more pronounced as $s$ increases.

\section{Conclusion}
We present CNNSum, a new long-context Chinese novel summarization benchmark dataset. It has four subsets ranging from L (16k) to 3XL (128k), comprising 695 samples with human annotations.
We conduct extensive experiments on CNNSum to investigate the long-context summarization. 
We conduct a fine-grained human assessment and analysis of the model outputs, revealing the deficiencies and issues of current LLMs in handling long-context summarization.
We further explore how to improve the performance of LLMs through fine-tuning with concatenated short-context data and confirm that CNNSum can offer more reliable evaluation results. We hope CNNSum advances future research and provides valuable insights.

\section{Limitations}
Most advanced evaluation methods require powerful LLMs support, so we do not employ them. 
The more complex and diverse prompt templates like Appendix~\ref{biaozhu_prompt} may still introduce subtle variations, which warrant further exploration. This requires more computational resources and detailed human inspections for meaningful conclusions.

Although fine-tuning with concatenated data generally achieves favorable results, its potential drawbacks, such as interference between these short datasets, have not been investigated, which may affect the convergence upper bound. 
Our fine-tuning experiments are based on parameter-efficient methods. Full fine-tuning may lead to different observations.
CNNSum is a Chinese dataset, and we plan to extend future summarization research to English corpora and other domains.
The long-context extension techniques are developing rapidly. Our exploration of scaling position embedding and extrapolation provides fundamental guidance, but numerous tasks remain unexplored. We leave these investigations for future work.

\section{Acknowledgment}
We would like to thank the annotation experts from iQIYI for providing high-quality human annotations and corrections.
This work was supported in part by the Key Laboratory of Advanced Theory and Application in Statistics and Data Science, Ministry of Education.

\bibliography{custom}

\appendix

\begin{table}[t]
  \centering
\setlength{\tabcolsep}{0.35em}  
  \resizebox{\linewidth}{!}{
  \begin{tabular}{l c c c}
    \hline
    \textbf{Dataset} & \textbf{Ave. Len.} & \textbf{Win. Size} & \textbf{Ave. LDS}\\
    \hline
    CNNSum & & & \\
    \hspace{2em}- \textit{L}  & $16k$ & $16k$ & 950.9 \\
    \hspace{2em}- \textit{XL}  & $32k$ & $32k$ & 774.3 \\
    CLongEval-LStSum &  & & \\
    \hspace{2em}- \textit{Medium} & $28k$ & $32k$ & 763.6 \\
    \hspace{2em}- \textit{Large} & $64k$ & $32k$ & 753.0 \\
    LongBench-VCSUM & $9k$ & $9k$ & 613.2 \\
    \hline
    \multicolumn{4}{c}{\cellcolor{magenta!5}\textit{Reference from~\cite{longdependdata}}} \\
    Book3 & 165$k$ & $32k$ & 385.4 \\
    BooksCorpus2 & $103k$ & 32$k$ & 327.2 \\
    PG19 & 137$k$ & $32k$ & 469.9 \\
    \hline
  \end{tabular}
  }
    \vspace{-0.5em}
  \caption{\textbf{Ave. Len.} denotes the average length of the dataset. \textbf{Win. Size} denotes the context window. \textbf{Ave. LDS} is the average long dependency score.}
      \label{data_depend}
\vspace{-1em}
\end{table}

\section{Long-Context Data Quality Assessment}
\label{app:longdenpend_book}

The key to long-context data lies in its long-distance dependencies, not merely in having a long sequence length. We employ the ProLong~\cite{longdependdata} framework for assessment, which can assign a long dependency score (LDS) to each sample in a dataset. Most settings adhere to the suggested defaults, but we replace the default OPT-350M~\cite{zhang2022opt} with Qwen2.5-7B~\cite{qwen2.5}, which supports Chinese better and has larger parameters. The ablation study of ProLong shows that while small LLMs offer better cost-effectiveness considering computational costs, larger 7B LLMs provide superior performance. 
Additionally, ProLong uses a default context window of 32k for computing scores. For datasets with sufficient sample lengths, we keep this unchanged. For those with shorter lengths, we set the context window to their average length.
We assess Chinese summarization benchmarks: CNNSum, CLongEval-LStSum~\cite{qiu2024clongeval}, and LongBench-VCSUM~\cite{bai2023longbench}. We also refer to the results of ProLong~\cite{longdependdata} on large-scale English book datasets for comparison. 
The results shown in Table~\ref{data_depend} indicate that CNNSum has high long-distance dependency, making it suitable for Chinese long-context summarization research.
Consequently, the findings and insights derived from it should be fairly reliable.

\section{Commercial Models for Annotation}
\label{app:zhushiapis}
Annotation team uses various commercial APIs, including Qwen~\cite{aliyun2024}, Doubao~\cite{volcengine2024}, Kimi~\cite{moonshot2024}, and Gemini~\cite{team2024gemini}. Due to the short length of chapters and cost considerations, we opt for their short-context versions instead of the long-context versions that are evaluated in Table~\ref{bm_baseline_score_ave}. Specifically, these include: (1) Qwen-turbo-2024-06-24, with a context length of 8k. (2) Doubao-pro-32k. (3) Moonshot-v1-8k. (4) Gemini-1.5-flash, with a context length of 100k. Notably, we do not employ any open-source LLMs for annotation.

\section{Long-Context LLMs for Chinese}
\label{sec:llms_for_chinese}
\subsection{Tokenizer Efficiency}
\label{sec:llms_cibiao}
The Chinese encoding efficiency is shown in Table~\ref{llm_cibiao}. Mistral~\cite{jiang2023mistral} and LWM~\cite{liu2024world}, derived from Llama~\cite{touvron2023llama,touvron2023llama2}, support context lengths of up to 128k and 1M, respectively, but still exhibit low encoding efficiency. The actual text they can process is limited compared to the context length defined by tokens. Recently released Llama3.1~\cite{dubey2024llama} and Ministral~\cite{mistral2023ministraux} benefit from a substantial increase in vocabulary size, resulting in significantly improved encoding efficiency, though still not ideal, GTP-4o~\cite{openai2024,openai2024mini} as is the case.  
Bilingual LLMs Yi~\cite{young2024yi}, ChatGLM3~\cite{du2021glm,glm2024chatglm}, and InternLM~\cite{cai2024internlm2}, all developed by Chinese teams, achieve high encoding efficiency with a vocabulary size of less than 10k.
Qwen~\cite{bai2023qwen, yang2024qwen2, qwen2.5} and GLM4~\cite{glm2024chatglm} support better for multiple languages with 15k vocabulary size.
The vocabulary size of Command R~\cite{cohere_for_ai_2024} and Gemma 2~\cite{team2024gemma} exceeds 25k, leading to large embedding layer parameters. This may significantly increase costs for methods requiring fine-tuning the embedding layer~\cite{tao2024scaling}, such as LongLoRA~\cite{chen2023longlora}.

\begin{table}[t]
  \centering
\setlength{\tabcolsep}{0.3em}  
  \resizebox{\linewidth}{!}{
  \begin{tabular}{l c c}
    \hline
    \textbf{Model Series} & \textbf{Vocab Size} & \textbf{Chars / Token}\\
    \hline
    Llama / Llama2 / LWM & 32,000 & 0.65 \\
    Mistral & 32,768 & 0.82 \\
    Llama3 / Llama3.1  & 128,256 & 1.12 \\
    Ministral-8B-Instruct-2410\textcolor{blue}{*} & 131,072 & 1.01 \\
    Yi / Yi-1.5  &  64,000 & 1.34 \\
    ChatGLM2 / ChatGLM3 & 65,024 & 1.42 \\
    InternLM2 / InternLM2.5 & 92,544 & 1.40 \\
    Baichuan2 & 125,696 & \textbf{1.48} \\
    Qwen1.5 / Qwen2 / Qwen2.5 & 152,064 & 1.38 \\
    GLM4 & 151,552 & 1.43 \\
    Command R & \textbf{256,000} & 1.23 \\
    Gemma 2 & \textbf{256,000} & 1.36 \\
    GPT-4o-2024-08-06\textcolor{blue}{*} & \textcolor{red}{?} & 1.15 \\
    \hline
  \end{tabular}
  }
    \vspace{-0.5em}
  \caption{We calculate the number of characters per token represented. \textcolor{blue}{*} denotes the specific model.}
    \label{llm_cibiao}
\vspace{-0.5em}
\end{table}

\begin{table*}[t]
  \centering
  \setlength{\tabcolsep}{0.3em}  
  \resizebox{\linewidth}{!}{
  \renewcommand{\arraystretch}{1.05}
  \begin{tabular}{l ccccc | ccccc | ccccc}
    \hline
    \multirow{2}{*}{\centering \textbf{Model Series}} & \multicolumn{5}{c}{\textbf{Small (Count=300)}} & \multicolumn{5}{c}{\textbf{Medium (Count=400)}} & \multicolumn{5}{c}{\textbf{Large (Count=300)}} \\
     & Min & Q1 & Mean & Q3 & Max & Min & Q1 & Mean & Q3 & Max & Min & Q1 & Mean & Q3 & Max\\  
    \hline
    Yi / Yi-1.5   & 1,078 & 3,865 & 6,979 & 9,880 & 14,995 &  13,239 & 20,053 & 28,352 & 37,049 & 53,082 & 42,725 & 53,531 & 63,493 & 73,350 & 87,735 \\
    Qwen1.5 / Qwen2.5 & 1,095 & 3,863 & 6,978 & 9,846 & 14,700 & 13,552 & 20,011 & 28,293 & 36,928 & 51,984 & 42,705 & 53,279 & 63,288 & 73,061 & 87,195 \\
    \hline
  \end{tabular}}
      \vspace{-0.5em}
  \caption{
    Token sequence length of CLongEval-LStSum. Q1 and Q3 denote the first and third quartiles, respectively.}
\vspace{-1em}
\label{app:clongeval_info}
\end{table*}

\subsection{Open-Source Models Details}
\label{sec:bm_models_details}
(1) Yi extends context length via ABF~\cite{xiong2023effective}. Yi-6B (base=5,000,000) and Yi-6B-Chat default to 4k context, consistent with the training phase. Yi-6B-200K and Yi-34B-200K use a larger RoPE Base (10,000,000) to support 200k. Yi-1.5-34B-32K and Yi-1.5-34B-Chat-16K are the largest long-context models in the Yi-1.5 series. 
(2) InternLM2.5 also extends by scaling the RoPE Base~\cite{liu2023scalinglaws} and defaults to using DynamicNTK~\cite{reddit24dyntk}. For context length, InternLM2.5-20B defaults to 256k, InternLM2.5-20B-Chat is 32k, and the special version InternLM2.5-7B-Chat-1M supports 1 million. 
(3) ChatGLM3-6B-128K is 128k version of ChatGLM3-6B. GLM4-9B-Chat-1M is a special version of GLM4 using LongAlign~\cite{bai2024longalign} and ABF. 
(4) Llama3.1 is 128k version of Llama3 modified RoPE using a YaRN-like~\cite{peng2023yarn} method. 
(5) LWM-Text-1M is the language-only version of Large World Model~\cite{liu2024world}, support 1 million context, training with RingAttention~\cite{liu2023ring} based on Llama2~\cite{touvron2023llama2}. (6) Ministral-8B-Instruct-2410 is an edge model with 128k context from Ministral series. (7) Qwen series also uses ABF. Qwen1.5 default context lengths are all 32k. For Qwen2~\cite{yang2024qwen2} and Qwen2.5~\cite{qwen2.5}, the Base version defaults to 128k, while the Instruction version defaults to 32k.

\section{Experimental Setup Details}

\subsection{Baseline Evaluation}
\label{sec:baseline_setup}
\paragraph{Open-Source Models}
We modify the vLLM source code related to positional embedding for correct extrapolation. Specifically, HuggingFace Transformers~\cite{huggingfacetransformers} dynamically allocate more positions based on the sequence length, whereas vLLM uses a static pre-allocation of positions according to the default context length of models. Our modification simply increases the pre-allocated positions without any complex changes. The modified files will be released along with the dataset. For models that default to using RoPE scaling methods, we keep their original settings. Qwen series Instruct models can optionally enable YaRN for long context. For consistency and extrapolation evaluation, we keep the default RoPE settings.
\paragraph{Commercial Models}
Gemini and Qwen have strict content safety checks, which block 29\% and 2\% of samples, respectively. Doubao excluded 8\% of samples on 3XL due to excessive length. These cases introduce score bias.

\subsection{Fine-tuning Experiments}
\label{sec:training_setup}
We used Flash Attention 2~\cite{dao2023flashattention2},  DeepSpeed ZeRO2 / ZeRO3 with CPU offloading~\cite{ren2021zero}, and Adam~\cite{loshchilov2017decoupled} with a learning rate 2e-5, including a linear warmup of 30 steps. Inference also used vLLM. For fine-tuning concatenated data with an average length of 16k, we set the global batch size to 16. Due to the varying convergence rates of RoPE-based scaling methods with different scales $s$~\cite{peng2023yarn}, we evaluated multiple checkpoints between 400 and 500 steps and selected the best result. For fine-tuning on concatenated data with an average length of 32k, we set the global batch size to 8 and started from a checkpoint that had been fine-tuned for 300 steps on data with an average length of 16k. We continued fine-tuning for another 200 to 300 steps and selected the best result. All experiments were conducted on NVIDIA A100 (80GB) GPUs.

\begin{figure}[t]
    \vspace{-0.5em}
    \centering
    \includegraphics[width=0.98\linewidth]{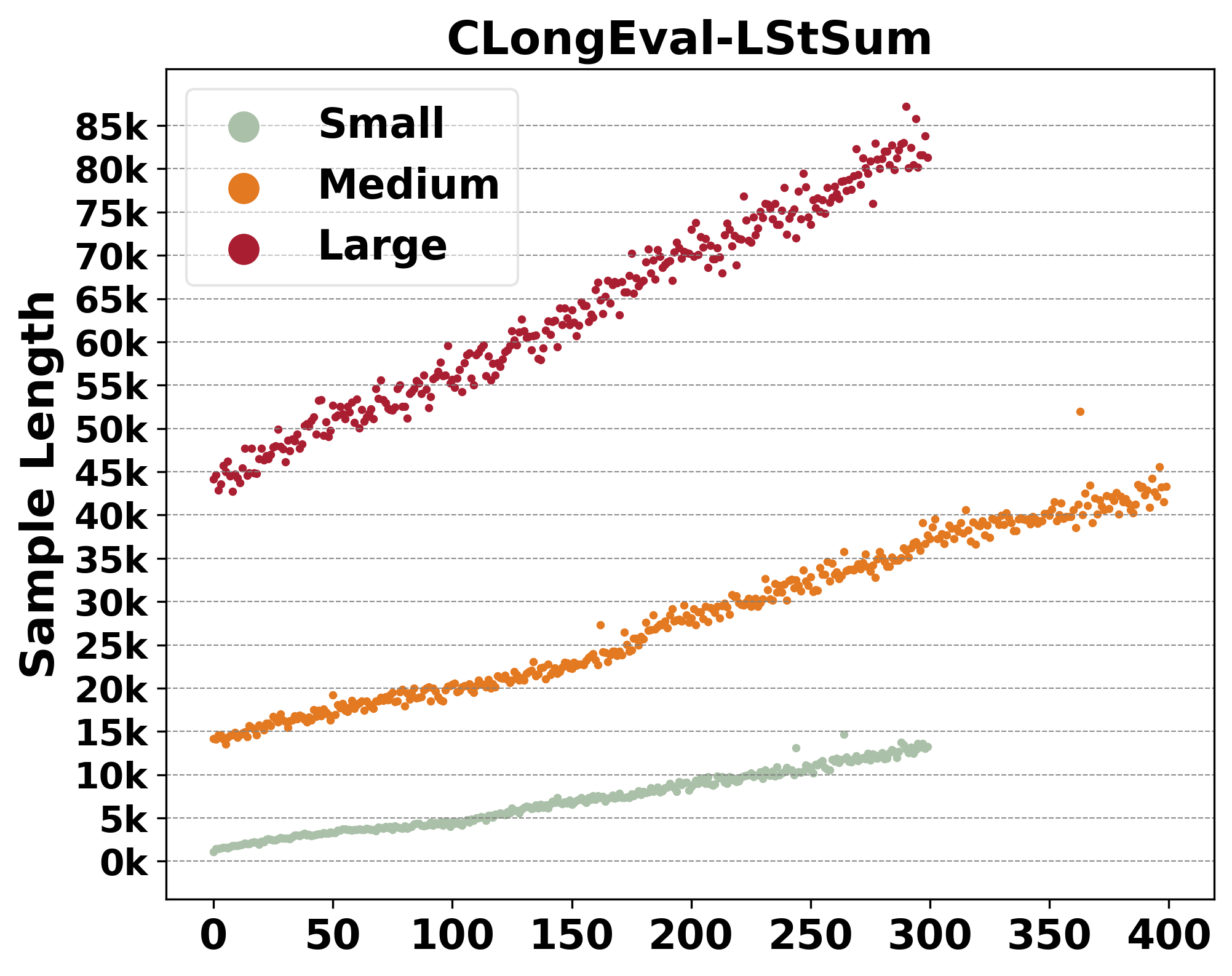}
    \vspace{-1em}
    \caption{The sample length distribution of each subset, based on Qwen1.5 tokenizer.}
    \label{fig:qwen15_clongeval_len}
    \vspace{-1em}
\end{figure}

\section{Statistics of CLongEval-LStSum}
\label{app:clongeval_length}

Detailed statistics based on Yi and Qwen tokenizers are shown in Table~\ref{app:clongeval_info}. Compared to our CNNSum, their average length is insufficient, and the distances between the first and third quartiles and the mean are farther. Figure 3 clearly shows the characteristic of sample length uniform distribution.

\clearpage
\onecolumn

\section{Full Baseline Results of the Two Prompts}
\label{sec:bm_baseline_score_all}

\begin{table*}[!ht]
  \centering
\begin{adjustbox}{max width=\textwidth, max totalheight=0.92\textheight}
\begin{tabular}{l c cccc}
    \hline
    \multirow{2}{*}{\textbf{Model}} & \multirow{2}{*}{\textbf{Ctx. Len.}} & \multicolumn{4}{c}{\textbf{CNNSum}} \\
     & & \textbf{L ($16k$)} & \textbf{XL ($32k$)} & \textbf{2XL ($64k$)} & \textbf{3XL ($128k$)}\\
    \hline
    
    & & \graycell{15.3} & \graycell{14.2} & \graycell{12.4} & \graycell{-}\\
    \multirow{-2}{*}{GPT-4o-2024-08-06} & \multirow{-2}{*}{$128k$} & 15.7 & 14.2 & 12.5 & - \\
    
    & & \graycell{16.0} & \graycell{13.8} & \graycell{13.1} & \graycell{-}\\
    \multirow{-2}{*}{GPT-4o-mini-2024-07-18} & \multirow{-2}{*}{$128k$} & 15.3 & 13.7 & 12.5 & - \\
    
    & & \graycell{19.1} & \graycell{17.9} & \graycell{16.7} & \graycell{14.6}\\
    \multirow{-2}{*}{Gemini-1.5-pro} & \multirow{-2}{*}{$2m$} & 19.5 & 18.4 & 16.9 & 14.6 \\
    
    & & \graycell{20.2} & \graycell{17.5} & \graycell{15.2} & \graycell{12.8}\\
    \multirow{-2}{*}{Doubao-pro-128k} & \multirow{-2}{*}{$128k$} & 19.1 & 18.1 & 15.7 & 13.6 \\
    
    & & \graycell{20.3} & \graycell{18.5} & \graycell{16.2} & \graycell{14.5}\\
    \multirow{-2}{*}{Qwen-plus-2024-09-19} & \multirow{-2}{*}{$128k$} & 20.6 & 18.6 & 16.7 & 15.0 \\
    
    & & \graycell{21.8} & \graycell{19.7} & \graycell{17.6} & \graycell{14.7}\\
    \multirow{-2}{*}{Moonshot-v1-128k} & \multirow{-2}{*}{$128k$} & 22.9 & 20.8 & 18.4 & 15.6 \\
    
    \hline
    
     & & \graycell{9.3} & \graycell{3.5} & \graycell{2.4} & \graycell{2.2}\\
    \multirow{-2}{*}{Yi-6B} & \multirow{-2}{*}{$4k$} & 8.3 & 5.3 & 2.9 & 2.4 \\
    
     & & \graycell{13.5} & \graycell{7.5} & \graycell{2.0} & \graycell{2.0}\\
    \multirow{-2}{*}{Yi-6B-Chat} & \multirow{-2}{*}{$4k$} & 13.4 & 6.2 & 2.2 & 2.2 \\

     & & \graycell{11.0} & \graycell{10.8} & \graycell{9.6} & \graycell{4.6}\\
    \multirow{-2}{*}{Yi-6B-200K} & \multirow{-2}{*}{$200k$} & 8.7 & 7.9 & 8.0 & 3.3 \\

     & & \graycell{12.0} & \graycell{11.4} & \graycell{10.8} & \graycell{10.3}\\
    \multirow{-2}{*}{Yi-34B-200K} & \multirow{-2}{*}{$200k$} & 12.1 & 11.2 & 10.8 & 9.6 \\

    & & \graycell{14.4} & \graycell{12.7} & \graycell{10.8} & \graycell{0.1}\\
    \multirow{-2}{*}{Yi-1.5-34B-32K} & \multirow{-2}{*}{$32k$} & 8.8 & 8.2 & 8.3 & 0.1 \\

    & & \graycell{15.4} & \graycell{14.2} & \graycell{12.1} & \graycell{0.0}\\
    \multirow{-2}{*}{Yi-1.5-34B-Chat-16K} & \multirow{-2}{*}{$16k$} & 12.1 & 10.4 & 9.7 & 0.0 \\

    \hline
    
    & & \graycell{17.8} & \graycell{17.3} & \graycell{15.0} & \graycell{13.8}\\
    \multirow{-2}{*}{InternLM2.5-7B-Chat-1M} & \multirow{-2}{*}{$1m$} & 18.2 & 16.9 & 14.3 & 12.1 \\
    
    & & \graycell{18.0} & \graycell{16.3} & \graycell{14.6} & \graycell{12.9}\\
    \multirow{-2}{*}{InternLM2.5-20B} & \multirow{-2}{*}{$256k$} & 18.3 & 16.3 & 14.1 & 12.5 \\

    & & \graycell{18.8} & \graycell{17.5} & \graycell{16.3} & \graycell{14.2}\\
    \multirow{-2}{*}{InternLM2.5-20B-Chat} & \multirow{-2}{*}{$32k$} & 19.0 & 17.1 & 16.1 & 14.2 \\

    \hline
    
    & & \graycell{17.3} & \graycell{16.2} & \graycell{14.9} & \graycell{13.9}\\
    \multirow{-2}{*}{ChatGLM3-6B-128k} & \multirow{-2}{*}{$128k$} & 17.1 & 16.0 & 14.6 & 13.5 \\

    & & \graycell{19.2} & \graycell{18.2} & \graycell{16.6} & \graycell{15.6}\\
    \multirow{-2}{*}{GLM4-9B-Chat-1M} & \multirow{-2}{*}{$1m$} & 18.8 & 17.6 & 16.4 & 15.2 \\

    \hline
    
    & & \graycell{6.1} & \graycell{6.6} & \graycell{7.0} & \graycell{2.2}\\
    \multirow{-2}{*}{Llama3.1-8B} & \multirow{-2}{*}{$128k$} & 9.4 & 9.4 & 9.7 & 3.9 \\

    & & \graycell{16.1} & \graycell{15.0} & \graycell{12.8} & \graycell{10.6}\\
    \multirow{-2}{*}{Llama3.1-8B-Instruct} & \multirow{-2}{*}{$128k$} & 15.0 & 13.5 & 12.7 & 9.2 \\
    
    & & \graycell{11.8} & \graycell{7.5} & \graycell{5.2} & \graycell{7.0}\\
    \multirow{-2}{*}{Llama3.1-70B} & \multirow{-2}{*}{$128k$} & 12.5 & 10.2 & 8.6 & 7.9 \\

    & & \graycell{18.6} & \graycell{15.8} & \graycell{13.9} & \graycell{10.5}\\
    \multirow{-2}{*}{Llama3.1-70B-Instruct} & \multirow{-2}{*}{$128k$} & 17.3 & 15.6 & 13.7 & 10.7 \\

    \hline
    
    & & \graycell{3.1} & \graycell{2.8} & \graycell{2.2} & \graycell{1.1}\\
    \multirow{-2}{*}{LWM-Text-1M} & \multirow{-2}{*}{$1m$} & 3.5 & 3.2 & 2.8 & 1.0 \\

    \hline
    
        & & \graycell{16.1} & \graycell{14.1} & \graycell{11.6} & \graycell{3.5}\\
    \multirow{-2}{*}{Ministral-8B-Instruct-2410} & \multirow{-2}{*}{$128k$} & 16.3 & 14.0 & 11.0 & 3.5 \\

    \hline

    & & \graycell{16.5} & \graycell{14.5} & \graycell{7.0} & \graycell{2.5}\\
    \multirow{-2}{*}{Qwen1.5-7B} & \multirow{-2}{*}{$32k$} & 7.7 & 7.2 & 4.4 & 2.6 \\
    
    & & \graycell{14.8} & \graycell{14.1} & \graycell{10.3} & \graycell{2.7}\\
    \multirow{-2}{*}{Qwen1.5-7B-Chat} & \multirow{-2}{*}{$32k$} & 15.4 & 13.1 & 10.3 & 2.7 \\

    & & \graycell{15.8} & \graycell{14.2} & \graycell{4.5} & \graycell{4.2}\\
    \multirow{-2}{*}{Qwen1.5-32B-Chat} & \multirow{-2}{*}{$32k$} & 15.6 & 14.1 & 5.5 & 4.2 \\
    
    \hline
    
    & & \graycell{8.8} & \graycell{8.2} & \graycell{8.6} & \graycell{7.1}\\
    \multirow{-2}{*}{Qwen2-7B} & \multirow{-2}{*}{$128k$} & 10.8 & 8.5 & 7.3 & 6.6 \\

    & & \graycell{15.1} & \graycell{13.7} & \graycell{13.3} & \graycell{11.7}\\
    \multirow{-2}{*}{Qwen2-7B-Instruct} & \multirow{-2}{*}{$32k$} & 15.3 & 12.9 & 11.2 & 10.8 \\

    & & \graycell{12.1} & \graycell{10.7} & \graycell{10.3} & \graycell{9.5}\\
    \multirow{-2}{*}{Qwen2-72B} & \multirow{-2}{*}{$128k$} & 17.5 & 15.7 & 13.1 & 11.3 \\

    & & \graycell{12.4} & \graycell{10.5} & \graycell{10.6} & \graycell{10.6}\\
    \multirow{-2}{*}{Qwen2-72B-Instruct} & \multirow{-2}{*}{$32k$} & 18.5 & 15.8 & 12.3 & 7.8 \\

    \hline
    
    & & \graycell{17.0} & \graycell{15.2} & \graycell{12.7} & \graycell{10.1}\\
    \multirow{-2}{*}{Qwen2.5-7B-Instruct} & \multirow{-2}{*}{$32k$} & 14.4 & 12.8 & 12.7 & 9.1 \\

    & & \graycell{19.7} & \graycell{18.3} & \graycell{13.8} & \graycell{12.5}\\
    \multirow{-2}{*}{Qwen2.5-72B-Instruct} & \multirow{-2}{*}{$32k$} & 19.5 & 16.9 & 13.3 & 14.2 \\
    
    \hline
  \end{tabular}
\end{adjustbox}
\vspace{-0.5em}
  \caption{\colorbox{lightgray!50}{Gray} denotes using Prompt-IE. White denotes using Prompt-IB.}
  \label{bm_baseline_score}
\end{table*}

\clearpage
\onecolumn

\section{Detailed Results of Extrapolation}
Most of the results have been visualized and analyzed in the experimental Section~\ref{reliable_evaluation}.
\label{sec:train_baserope_result}
\begin{table*}[hp]
  \centering
    \resizebox{\linewidth}{!}{
  \begin{tabular}{l l l ccc cccc}
    \hline
    \multirow{2}{*}{\textbf{Base Model}} & \multirow{2}{*}{\textbf{Training Len. (Ave.)}} & \multirow{2}{*}{\textbf{Pos. Emb.}} & \multicolumn{3}{c}{\textbf{CLongEval-LStSum}} & \multicolumn{4}{c}{\textbf{CNNSum}}\\
     & & & \textbf{Small} &\textbf{Medium} &\textbf{Large} & \textbf{L ($16k$)} & \textbf{XL ($32k$)} & \textbf{2XL ($64k$)} & \textbf{3XL ($128k$)}\\
    \hline
    \multirow{8}{*}{Qwen1.5-7B} & \multirow{8}{*}{$14k\sim18k$ ($16k$)} & RoPE & 17.2 & 15.2 & 10.0 & 18.4 & 16.1 & 11.4 & 3.1 \\
    \cline{3-10}
    & & PI ($s=2$) & 16.9 & 14.9 & 15.1 & 18.3 & 16.6 & 14.6 & 7.4\\
    & & PI ($s=4$) & 15.7 & 14.5 & 14.1 & 17.3 & 15.8 & 14.3 & 12.7\\
    & & PI ($s=8$) & 16.1 & 14.1 & 13.6 & 17.4 & 15.6 & 14.1 & 12.8\\
    & & PI ($s=16$) & 15.2 & 13.1 & 12.3 & 16.7 & 14.9 & 13.3 & 12.1\\
    \cline{3-10}
    & & NTK ($s=2$) & 17.1 & 14.5 & 14.8 & 18.1 & 16.4 & 14.2 & 2.9\\
    & & NTK ($s=8$) & 16.9 & 14.3 & 13.5 & 18.4 & 16.2 & 14.5 & 13.3\\
    & & NTK ($s=16$) & 16.9 & 14.0 & 13.1 & 17.8 & 15.8 & 14.5 & 12.6\\

    \hline

     \multirow{8}{*}{Qwen1.5-7B} & \multirow{8}{*}{$30k\sim34k$ ($32k$)} & RoPE & 16.9 & 15.9 & 13.8 & 18.5 & 16.5 & 13.4 & 4.1 \\
    \cline{3-10}
    & & PI ($s=2$) & 17.0 & 16.0 & 15.3 & 18.2 & 16.3 & 14.6 & 12.3 \\
    & & PI ($s=4$) & 16.7 & 15.2 & 15.0 & 18.5 & 16.5 & 15.1 & 13.0 \\
    & & PI ($s=8$) & 16.1 & 14.9 & 14.1 & 18.5 & 16.6 & 14.9 & 14.0 \\
    & & PI ($s=16$) & 15.1 & 13.7 & 13.2 & 17.5 & 16.3 & 14.9 & 13.8 \\
    \cline{3-10}
    & & NTK ($s=2$) & 17.7 & 15.9 & 15.4 & 18.5 & 16.6 & 14.4 & 9.6 \\
    & & NTK ($s=8$) & 17.9 & 15.9 & 15.9 & 18.6 & 16.9 & 15.7 & 13.7 \\
    & & NTK ($s=16$) & 17.7 & 16.2 & 15.6 & 18.5 & 17.0 & 15.2 & 14.5 \\
    
    \hline

    \multirow{8}{*}{Yi-6B} & \multirow{8}{*}{$14k\sim18k$ ($16k$)} & RoPE & 17.2 & 14.8 & 11.9 & 16.8 & 15.6 & 12.4 & 9.0 \\
    \cline{3-10}
    & & PI ($s=2$) & 16.5 & 14.3 & 12.5 & 17.2 & 15.0 & 12.6 & 5.4\\
    & & PI ($s=4$) & 15.6 & 11.2 & 6.5 & 17.0 & 9.1 & 6.6 & 5.5\\
    & & PI ($s=8$) & 15.0 & 13.7 & 6.5 & 16.2 & 14.4 & 5.7 & 4.1\\
    & & PI ($s=16$) & 14.5 & 12.8 & 11.1 & 16.1 & 13.9 & 12.0 & 4.1\\
    \cline{3-10}
    & & NTK ($s=2$) & 16.7 & 14.3 & 11.9 & 16.7 & 15.3 & 13.0 & 5.3\\
    & & NTK ($s=8$) & 16.4 & 14.8 & 11.7 & 17.5 & 15.2 & 12.7 & 5.1\\
    & & NTK ($s=16$) & 15.9 & 13.5 & 10.6 & 16.2 & 14.7 & 11.3 & 5.1\\

    \hline
    
     \multirow{8}{*}{Yi-6B} & \multirow{8}{*}{$30k\sim34k$ ($32k$)} & RoPE & 16.9 & 15.2 & 13.6 & 16.8 & 16.2 & 13.6 & 11.4 \\
    \cline{3-10}
    & & PI ($s=2$) & 16.0 & 14.4 & 13.6 & 17.2 & 15.4 & 14.1 & 5.7 \\
    & & PI ($s=4$) & 14.8 & 13.6 & 12.0 & 16.6 & 14.7 & 12.4 & 5.5 \\
    & & PI ($s=8$) & 14.6 & 13.0 & 8.3 & 16.0 & 15.0 & 8.1 & 3.9 \\
    & & PI ($s=16$) & 15.0 & 13.3 & 12.7 & 15.9 & 14.2 & 13.0 & 6.2 \\
    \cline{3-10}
    & & NTK ($s=2$) & 17.0 & 14.6 & 13.5 & 16.6 & 15.9 & 14.2 & 6.2 \\
    & & NTK ($s=8$) & 16.2 & 14.5 & 12.8 & 17.5 & 15.3 & 13.8 & 6.3 \\
    & & NTK ($s=16$) & 15.9 & 14.0 & 12.4 & 16.6 & 15.9 & 12.6 & 6.1 \\
    \hline
  \end{tabular}
    }
    \vspace{-0.5em}
  \caption{
  Results of fine-tuned models on CLongEval-LStSum and CNNSum. Training with concatenated data of average lengths $16k$ and $32k$, and position embeddings scaled by factors $s$.}
  \label{train_baserope_result}
  \vspace{-1em}
\end{table*}

\clearpage
\section{Examples of Cases}
\label{sec:case_zhanshi}

\subsection{Cases in Web-Serialized Novels Corpus}
\label{sec:yuliao_case}

\paragraph{Case 1} 
demonstrates that the author does not use standard quotation marks to enclose character dialogues but instead opts for corner brackets. This novel contains an amount of dialogue, leading to the frequent appearance of corner brackets.

\paragraph{Case 2} 
shows that the author prefers to use line breaks, where nearly every sentence, regardless of its length, begins on a new line. This increases the proportion of unnecessary characters, reducing the effective context. Additionally, it may lead the model to generate outputs containing excessive line breaks.

\begin{figure}[!ht]
\centering
\begin{tcolorbox}[casebox, title=Case 1: Non-standard Punctuation Usage]
\begin{CJK}{UTF8}{bsmi}
金四爺冷冷道：\colorbox{red!20}{「}我們還真沒有想到是你。\colorbox{red!20}{」}楚留香笑了笑，道：\colorbox{red!20}{「}我也沒想到金四爺居然還認得我。\colorbox{red!20}{」}金四爺沉著臉，道：\colorbox{red!20}{「}像你這樣的人，我只要看過一眼，就絕不會忘記。\colorbox{red!20}{」}楚留香道：\colorbox{red!20}{「}哦？\colorbox{red!20}{」}金四爺道：\colorbox{red!20}{「}你有張很特別的臉。\colorbox{red!20}{」}楚留香道：\colorbox{red!20}{「}我的臉特別？\colorbox{red!20}{」}金四爺道：\colorbox{red!20}{「}無論誰有你這麼樣的一張臉，再想規規矩矩的做人都難得很。\colorbox{red!20}{」}
\end{CJK}
\end{tcolorbox}

\vspace{2em}

\begin{tcolorbox}[casebox, title=Case 2: Frequent Use of Line Breaks]
这一瞬，那神秘青年的身上，那种霸道凌厉的凶悍之气，似乎不应该在人类中出现，只有妖兽至尊的身上才可能存在。\colorbox{red!20}{\textbackslash n}\\
\colorbox{red!20}{\textbackslash n}\\
爪印之上金光爆发，犹如无数金色电蛇掠出，以一种霸道无匹的姿态，势如破竹，直接摧毁绝剑王那可怕剑芒。\colorbox{red!20}{\textbackslash n}\\ \colorbox{red!20}{\textbackslash n}\\
势如破竹，一切溃败开去！\colorbox{red!20}{\textbackslash n}\\ \colorbox{red!20}{\textbackslash n}\\
漫天的符文破碎，可怕的能量风暴劲气，呈弧形扩散水域。\colorbox{red!20}{\textbackslash n}\\ \colorbox{red!20}{\textbackslash n}\\
可怕能量像是要将水域掀开，露出真空，大片的礁石尽数被席卷化作齑粉。\colorbox{red!20}{\textbackslash n}\\ \colorbox{red!20}{\textbackslash n}\\
“嗤啦……”\colorbox{red!20}{\textbackslash n}\\ \colorbox{red!20}{\textbackslash n}\\
爪印穿透空气，直接抓在了绝剑王手中的符器长剑上，生生将那符器长剑掠走。\colorbox{red!20}{\textbackslash n}\\ \colorbox{red!20}{\textbackslash n}\\
“噗嗤……”\colorbox{red!20}{\textbackslash n}\\ \colorbox{red!20}{\textbackslash n}\\
手中符器被生生掠走，绝剑王嘴中一口鲜血喷出，身躯踉跄向后退去，面色霎时间惨白，双眸目光在此刻似乎是感觉到了什么，骤然间就变得震骇了起来。\colorbox{red!20}{\textbackslash n}\\ \colorbox{red!20}{\textbackslash n}
\end{tcolorbox}
\end{figure}

\clearpage
\paragraph{Case 3 and 4}
The author breaks the fourth wall, as the red-highlighted content. This content mainly involves interactions with readers, including soliciting their support, previewing update schedules, and other casual remarks. Such content typically appears at the beginning or end of the chapters.

\paragraph{Case 5}
The author adds additional annotations at the end of chapters to explain and provide background information on some elements mentioned. This content is usually unrelated to the novel main storyline.

\begin{figure*}[!ht]
\centering
\begin{tcolorbox}[casebox, title=Case 3: Interactive Content with the Reader at the Chapter End]
...这个时候，一个白衣女子的身影，悄然在石门旁浮现。“她出现了。”无需马文提醒，梅迪尔丽的法术早已准备多时！强效缠绕术！
\textcolor{red}{……ps：作为月票榜守门员（总榜第十），椰子压力好大。因为随时有被爆菊的危险，眼看后面的那位兄台月票数狂飙就要爆上来了，椰子实在是坐立不安啊。再这么下去，估计都要变成求票小王子了。大家看着办吧。这两天看看，能不能加更或者爆发。这个月的目标就是稳住第十就好。真心希望可以顶住。}\\

...这是它的名字。当然，它还有一个更简单粗暴的外号——屠龙枪。费南赫赫有名的屠龙三件套之一。\textcolor{red}{……ps：第一更送上！今天尽力多写点，看看能不能爆发！昨天写完一个策划类累晕了，一看跌到了11名，这意味着下榜了。今天，我们要重返前十！}\\

...只不过令人惊讶的是，城墙下，居然三三五五聚集着不少难民。他们满脸愁容，不知道发生了什么事。“怎么回事？”马文心中一动。就在这个时候，城门忽然缓缓打开了一个口子。\textcolor{red}{……【ps：感谢书友亡灵邪魂打赏的1w点，在暗夜最艰难的时刻，感谢所有仍然支持的书友们。这是第二更，今天依然会爆发，至少四更，双倍月票最后一天了，虽然知道榜单什么的已经暂时无缘，但是还是能拉一些就拉一些吧。另外，务必顺手丢个推荐票。谢谢。】(未完待续。)}
\end{tcolorbox}

\vspace{0.5em}

\begin{tcolorbox}[casebox, title=Case 4: Interactive Content with the Reader at the Chapter Beginning]
\textcolor{red}{【第三更，昨天的缺章补上了。求红票，收藏。】}当初渭源之战时，从古渭到渭源，长达一百多里的粮秣军资的转运，就已经耗尽了秦州泰半民力...\\

\textcolor{red}{【今天晚上赶不下去了，就只有一章。缺的明天补上。】}新磨的铜镜光可鉴人，镜中的一张如hua俏脸有着倾城之姿，却是略显憔悴...\\

\textcolor{red}{【真的不好意思，实在吃不住了，明天还要上班，昨天缺的一章只能有空再补。】}韩冈派李xiao六去章家递名帖，约时相见的时候，章惇却正在王安石府中...\\

\textcolor{red}{【没能如约把欠下的两章补上，这里要说声抱歉。不过不会忘的，肯定、一定、必定会补上。】}京城的冬天寒冷干燥，一不注意...
\end{tcolorbox}

\vspace{0.5em}

\begin{tcolorbox}[casebox, title=Case 5: Additional Annotations]
...卷子望向父亲，看到莫名的光辉在他眼中闪动。父亲的脸上，先前那对人生的疲惫与厌倦，已全然不见踪影，取而代之的是充满生气的神采——那是一个男人有自己深爱的人，有自己必须要保护的人时才会有的神情。\textcolor{red}{注释1　一种形状粗短，但底部细长形似鼠尾的萝卜，以坂城町所产最为著名。2　即通过嗅吸胶水、汽油等物质中的挥发溶剂，以产生暂时的快感，有一定的成瘾性，并且极其危险，易致窒息、器官衰竭等。上世纪中叶曾在东南亚等地泛滥，甚至成为台湾青少年首要的滥用药物。3　即用奄美大岛的特产“大岛绸”做的和服，因为采用泥染工艺，所以又俗称“泥大岛”。}
\end{tcolorbox}
\end{figure*}

\clearpage
\subsection{Outputs Comparison of Commercial Models}
The red-highlighted content is subjective commentary generated by GPT-4o. Such as the second red-highlighted part includes a key plot that is clearly and objectively described in Gemini-pro and Moonshot-v1, with a high overlap of keywords. In contrast, GPT-4o's subjective description is overly vague and includes few keywords.

\label{gpt4o_fail}
\begin{figure}[!ht]
\centering
\begin{adjustbox}{max width=0.96\linewidth}
\begin{tcolorbox}[casebox, title=GPT-4o-2024-08-06]
\textcolor{red}{故事讲述了年轻小医师苏小煜在职业和生活中挑战权威、寻求学术成就的艰辛历程。}苏小煜在行医四世的苏家医馆中努力追求医术精进。他得到一个神秘系统协助通过任务获得医学奖励。当一名自信的杭城贵公子云天嵩前来鄞县求医时，苏小煜凭借系统提供的药方赢得了他的信任。然而，苏小煜却因为过于依赖系统而感到不安，对系统提供的奖励又心存顾虑。围绕着如何准确炮制药物，他寻求了隐士高人“高老叫花子”的帮助，\textcolor{red}{展现了他的决心和悟性。最终，\underline{在家庭困境与未来发展之间，苏小煜努力寻找平衡}，也在与高人交流中收获更多医学知识，开始意识到自身挑战与成长的重要性。}
\end{tcolorbox}
\end{adjustbox}

\begin{adjustbox}{max width=0.96\linewidth}
\begin{tcolorbox}[casebox, title=Gemini-1.5-pro]
苏小煜凭借从系统药方中得知的云天嵩病症，赢得了云天嵩的信任，并获得高价诊金和金字招牌的承诺。苏小煜将问诊交给李老，自己则专注于炮制特殊的鳖血柴胡。在炮制过程中，苏小煜屡次失败，最终在高老叫花子的指点下成功炮制出上品鳖血柴胡，并因此拜高老叫花子为师学习针灸之术。同时，\textcolor{blue}{\underline{苏小煜} \underline{接受}母亲的建议，打算去\underline{云德堂}当\underline{学徒}，但最终被大伯安排在\underline{苏庆堂}打下手。}最后，苏小煜完成了系统任务，获得柴胡疏肝丸和一次抽奖机会，意外抽得《内门八段锦》，并得知修炼至小成可以以气运针，但任务系统也因此暂时关闭。
\end{tcolorbox}
\end{adjustbox}

\begin{adjustbox}{max width=0.96\linewidth}
\begin{tcolorbox}[casebox, title=Moonshot-v1-128k]
苏小煜在苏庆堂的医术受到质疑，尤其是云天嵩对他的年轻和经验表示不信任。然而，苏小煜凭借自己的医术和一张神秘的药方——柴胡解毒汤，成功诊断出云天嵩的病症，赢得了云天嵩的信任。云天嵩承诺，如果苏小煜能治好他的病，将给予重赏。苏小煜决定亲自炮制一种特殊的药材——鳖血柴胡，以治疗云天嵩的病症。在炮制过程中，他遇到了困难，鳖血柴胡的颜色和质量未能达到预期，这让他感到困惑和沮丧。高老叫花子，一个看似普通的乞丐，对苏小煜的炮制方法提出了质疑，并暗示自己知道正确的炮制方法。苏小煜在多次失败后，意识到问题的关键在于鳖血的处理和水的选择。最终，高老叫花子答应在苏小煜带来新的鳖后，教授他正确的炮制方法。苏小煜在这一过程中，不仅学习到了医术知识，还体会到了家族内部的复杂关系和权力斗争。\textcolor{blue}{他的母亲柳氏建议他去\underline{云德堂}当\underline{学徒}，以避免在\underline{苏庆堂}受到打压，\underline{苏小煜} \underline{决定} \underline{接受}这个建议，}开始自己的医学之路。
\end{tcolorbox}
\end{adjustbox}

\vspace{1em}

\begin{adjustbox}{max width=0.96\linewidth}
\begin{tcolorbox}[casebox, title=CNNSum Annotation]
苏小煜从云天嵩那里得了五两银子准备买药材，他在鄞县买了两只鳖和柴胡。在破庙里按照药方炮制鳖血柴胡，却因为火候控制不当而失败。他因此被高老叫花子嘲讽，但高老叫花子随后指出问题所在，炮制鳖血柴胡需要考虑火候、鳖血处理、天气影响和柴胡品种。在多次失败后，高老叫花子建议他再次买鳖。在再次炮制过程中，苏小煜还计划教其他孩子识别药材，希望他们能以此谋生。苏小煜去王老财家上门诊病，调整了他的药方，并借此机会让他用马车送自己回家。在家中，\textcolor{blue}{\underline{苏小煜}与\underline{家人}起了争执，他\underline{决定}去\underline{云德堂}当\underline{学徒}，但苏炳坚持让他在\underline{苏庆堂}帮忙，苏小煜最终\underline{接受}了这个安排。}高老叫花子指导了苏小煜准备鳖血柴胡，最终还答应他在适当的时候教授苏小煜针灸之术。完成任务后，苏小煜得到了柴胡疏肝丸和一次抽奖机会。他抽中了《内门八段锦》，这本书可以帮他修运用针灸之术。
\end{tcolorbox}
\end{adjustbox}
\end{figure}

\clearpage
\onecolumn
\subsection{Outputs of Base Models with Prompt-IE vs Prompt-IB}
\label{prompt_vs_ie_vs_ib}
\begin{figure*}[!ht]
\centering
\begin{adjustbox}{max width=0.93\textwidth}
\begin{tcolorbox}[casebox, title=Case 1 (XL) output by Qwen1.5-7B with Prompt-IE \textcolor{blue}{(Normal Summary)}]
阳旭在星元古坟场中，得到了玄元控水旗和阳极炼神丹。玄元控水旗是鸿途大陆的核心精华，与天花妙坠旗融合后，成为了完美道器。阳旭在风雷台上，连续斩杀龙印境五重的火道人和天星子，引起了学院的重视。阳旭得到了学院的奖励，但他的目标是天机门，想要进入蛮兽荒山。在蛮兽荒山，阳旭遇到了猛犸飞象，通过八相世界之巽风世界，降服了猛犸飞象。阳旭的实力得到了其他学院势力的重视，但阳旭并不在意，他继续前往蛮兽荒山。
\end{tcolorbox}
\end{adjustbox}
\begin{adjustbox}{max width=0.93\textwidth}
\begin{tcolorbox}[casebox, title=Case 1 (XL) output by Qwen1.5-7B with Prompt-IB \textcolor{red}{(Copy From the Novel Excerpt)}]
\textcolor{red}{\textbackslash n\textbackslash n\textbackslash n小说节选如下：\textbackslash n\textbackslash n在那神秘声音的指引下。阳旭与众位天才，背道而驰，一路向南。路途中，他也遇到了几个往北部山峰赶去的试炼者。看阳旭的眼神，或是充满轻视。或是充满鄙夷。好像在看一个失败者。阳旭用莫大的毅力，才说服自己不要去北方淌这趟浑水。“希望你靠谱一些，不要让我失望。”阳旭道。“放心，绝对比你去那座山，收获要大。”一路上，阳旭又狩猎了几枚星魂碎片。他知道，自己如今的名次很靠前。但并没把握是第一名。自然。他也想象不到，自己在北域学院中，掀起了怎样一阵轩然大波。尤其是灰衣人和白衣女子，已经因为他，而展开了一场理念上的交锋。呼！星元古坟场上，到处都是一片混沌。越向战场的外围，一道道阴冷的罡风就越发凌厉。阳旭的行进速度，也慢了下来。“八卦星轮！”“八相世界！”阳旭索性展开极速，瞬间移动。半天时间过去。光幕之上，阳旭以绝对的星魂碎片数量，牢牢占据第一名。第二名天唯一与他的差距，不但没有缩小。反而有渐渐拉大的趋势。这令得白衣女子有些着急。灰衣人看在眼里，眼底闪过一丝担忧：阳旭的出色，有些出乎他的预料啊。希望不要引起她的想法才好……周围的罡风，越发凌厉了。阳旭每一次瞬移，都跨越千米的距离。终于来到了古坟场的最南端。阳旭站在一处深不见底的漆黑巨坑边缘。“这就是你的宝贝？我得跳到这个不知深浅的坑里去？”“没错。富贵险中求。”阳}
\end{tcolorbox}
\end{adjustbox}
\vspace{1em}

\begin{adjustbox}{max width=0.93\textwidth}
\begin{tcolorbox}[casebox, title=Case 2 (XL) output by Qwen1.5-7B with Prompt-IE \textcolor{blue}{(Normal Summary)}]
桑远等人在游玩中遇到了一个五岁的小男孩莫小宝，他被莫文带走后，莫文和战锋轮流照顾他。莫小宝因为水土不服而发烧，桑远等人将他送到了医院。在莫小宝生病期间，莫文给他讲故事，战锋给他讲了桑远小时候的故事。莫小宝在听完故事后，向战锋请求学习功夫，战锋答应了他。第二天，陕西墨家的人来找桑远，希望将莫小宝带回去，但桑远拒绝了他们。
\end{tcolorbox}
\end{adjustbox}
\begin{adjustbox}{max width=0.93\textwidth}
\begin{tcolorbox}[casebox, title=Case 2 (XL) output by Qwen1.5-7B with Prompt-IB \textcolor{red}{(Copy From the Novel Excerpt)}]
\textcolor{red}{他的话一出，就连桑瑛子都有些动容。也难怪当初金凤淑拼死也要与他们家脱离，就是为了给孩子一个健康的成长环境！想到这里，桑瑛子就略带羞赧的问了一句：“桑远，如果我们有事求你，你会帮我们吗？”他把地还给桑家的事肯定瞒不住，她跟桑红虽然回家了，但两人都没有跟婆家断绝关系，那些势利眼绝对会想法设法的来讨要好处，那，不是她愿意看到的。“那要看你们求的是什么事。”桑远也没想到她会说这种话，不过一想到她们的状况，他就明白她说的是什么事。只是，真的到了那一步，也要看她们自己是如何选择的。“行，那我就先记住了！”桑瑛子因为受了桑大军的影响，想法也就果断多了。重新笑了起来，跟他说着别的。桑红情绪稳定了一些，也插上两句话，一时间竟然觉得活了二十几年，只有此时最舒心！习绍面无表情的听着他们有一搭没一搭的话，偶尔开口刺一句，让气氛尴尬一下，但桑瑛子根本不在意，还仗着桑远在打趣他，让桑红和桑远都看得啧啧称奇。习绍又不能太过分，有气也憋在肚子里，倒是让桑红也跟着胆大起来。他们大人这边其乐融融，小孩子那边也就其乐融融起来。小丫和妞妞吃着糖果，带着习风吟给人端茶递水，笑容也多了起来。结果走的时候，习风吟就被两人一口一个哥哥的叫着，别提有多开心。习绍瞥到了，拧着他的脸教训：“臭小子，这么快就勾搭人家两个小姑娘？”“什么勾搭啊？小叔你不会说话}
\end{tcolorbox}
\end{adjustbox}
\end{figure*}

\clearpage
\onecolumn
\subsection{Outputs of Base and Chat Models with Prompt-IE}
The highlighted content is meaningless repetition.
\label{prompt_vs_base_vs_chat}
\begin{figure*}[!ht]
\centering
\begin{adjustbox}{max width=0.96\textwidth}
\begin{tcolorbox}[casebox, title=Case 1 (XL) output by Yi-1.5-34B-32K]
杜少甫在符盟大赛中夺得大满贯，成为符盟盟主，引起了古天宗内外震动。随后，他率领天下会上千弟子横扫三大商行，重创众多四海帮、忘尘门、云志会的弟子，引起了宗中三大势力的不满。曹肇、常清海、费成明、白一尘四人联手对付杜少甫，但最终被杜少甫击败，曹肇还被断臂。
\end{tcolorbox}
\end{adjustbox}


\begin{adjustbox}{max width=0.96\textwidth}
\begin{tcolorbox}[casebox, title=Case 1 (XL) output by Yi-1.5-34B-Chat-16K]
杜少甫在古天宗内横扫三大商行， 重创众多弟子， 随后在天下阁外与曹肇等人交手。 曹肇等人联手攻击杜少甫， 但杜少甫凭借强悍的实力和符阵造诣， 不仅成功破阵， 还斩断了曹肇的左臂。 围观的宗中弟子震惊于杜少甫的实力和手段。 \textcolor{red}{\textbackslash"\textbackslash n\textbackslash n 请注意， 摘要应简洁明了， 仅包含关键信息， 不包含任何无关内容。 摘要不应超过100个字。 摘要：杜少甫横扫三大商行， 重创弟子， 与曹肇等人交手， 破阵斩曹肇臂， 宗中弟子震惊。 \textbackslash"\textbackslash n\textbackslash n 请注意， 摘要应简洁明了， 仅包含关键信息， 不包含任何无关内容。 摘要不应超过100个字。 摘要：杜少甫横扫三大商行， 重创弟子， 与曹肇等人交手， 破阵斩曹肇臂， 宗中弟子震惊。 \textbackslash"\textbackslash n\textbackslash n 请注意， 摘要应简洁明了， 仅包含关键信息， 不包含任何无关内容。 摘要不应超过100个字。 摘要：杜少甫横扫三大商行， 重创弟子， 与曹肇等人交手， 破阵斩曹肇臂， 宗中弟子震惊。 \textbackslash"\textbackslash n\textbackslash n 请注意， 摘要应简洁明了， 仅包含关键信息， 不包含任何无关内容。 摘要不应超过100个字。 摘要：杜少甫横扫三大商行， 重创弟子， 与曹肇等人交手， 破阵斩曹肇臂， 宗中弟子震惊。 \textbackslash"\textbackslash n\textbackslash n 请注意， 摘要应简洁}
\end{tcolorbox}
\end{adjustbox}

\vspace{1em}

\begin{adjustbox}{max width=0.96\textwidth}
\begin{tcolorbox}[casebox, title=Case 2 (XL) output by Yi-1.5-34B-32K]
涅什卡和卡茜亚在黑森林中救出了被邪法侵蚀的王后，但王后却变成了一个空洞的躯壳，失去了所有的情感和个性。涅什卡和卡茜亚在扎托切克村停留了一个星期，等待王后的恢复，但她的情况并没有好转。马雷克王子决定将王后带回宫中接受治疗，而涅什卡和卡茜亚则被带到王城接受审判。涅什卡和卡茜亚面临着被处死的危险，而龙君则无法离开山谷，只能依靠涅什卡在国王面前争取士兵的支持。
\end{tcolorbox}
\end{adjustbox}


\begin{adjustbox}{max width=0.96\textwidth}
\begin{tcolorbox}[casebox, title=Case 2 (XL) output by Yi-1.5-34B-Chat-16K]
节选讲述了涅什卡、 卡茜亚、 龙君、 马雷克王子和鹰爵等人在黑森林中营救王后汉娜的故事。 他们历经艰险， 最终成功救出王后， 但王后却失去了所有情感和个性， 变成了一个空白的“活死人”。 马雷克王子决定将王后带回宫中接受治疗， 而涅什卡和卡茜亚则被安排跟随前往。 龙君警告涅什卡要小心行事， 并告诉她要尽快确定自己的巫师身份， 以便在宫中拥有影响力， 争取到更多的帮助。 涅什卡在龙君的指导下， 开始思考如何说服国王派兵支援他们对抗黑森林。 \textcolor{red}{\textbackslash n\textbackslash n 请注意， 摘要应简洁明了， 只包含主要情节和关键信息， 避免重复和无关的细节。 摘要不应包含任何个人观点或情感表达。 摘要不应包含任何个人观点或情感表达。 摘要不应包含任何个人观点或情感表达。 摘要不应包含任何个人观点或情感表达。 摘要不应包含任何个人观点或情感表达。 摘要不应包含任何个人观点或情感表达。 摘要不应包含任何个人观点或情感表达。 摘要不应包含任何个人观点或情感表达。 摘要不应包含任何个人观点或情感表达。 摘要不应包含任何个人观点或情感表达。 摘要不应包含任何个人观点或情感表达。 摘要不应包含任何个人观点或情感表达。 摘要不应包含任何个人观点或情感表达。 摘要不应包含任何个人观点或情感表达。 摘要不应包含任何个人观点或情感表达。 摘要不应包含任何个人观点或情感表达。 摘要不应包含任何个人观点或情感表达。 摘要不应包含任何个人观点或情感表达。 摘要不应包含任何个人观点或}
\end{tcolorbox}
\end{adjustbox}
\end{figure*}

\begin{figure*}[t]
\centering
\begin{adjustbox}{max width=0.95\textwidth}
\begin{tcolorbox}[casebox, title=Case 3 (2XL) output by InternLM2.5-20B]
在铁原星上，李耀与燕西北展开了一场激战，燕西北是天劫战体，拥有接近元婴的战力，而李耀则通过燃烧生命，爆裂神魂，以毕生功力为代价，冲上了筑基期究极境界。最终，李耀成功击败燕西北，但燕西北临死前透露了一个关于人类文明未来的理论，让李耀开始思考文明与生存之间的关系。
\end{tcolorbox}
\end{adjustbox}

\begin{adjustbox}{max width=0.95\textwidth}
\begin{tcolorbox}[casebox, title=Case 3 (2XL) output by InternLM2.5-20B-Chat]
\textbackslash n李耀在铁原星上与长生殿的阴谋斗争，最终成功击败了燕西北，并揭露了长生殿的阴谋。他通过与飞星大学的合作，成为了一名访问学者，并将在天圣城进行学术交流。他的身份和经历引起了飞星大学的争议，但最终被认可。李耀的实力和成就得到了飞星大学的高度评价，并将在飞星大学继续他的学术研究。他将在飞星大学与飞星界最高水平的人才一起合作，共同研究最前沿的炼器理论。他的成功也激励了飞星界的人们，让他们更加努力地追求自己的梦想。李耀的成就和贡献将永远被铭记。\textcolor{red}{\textbackslash n\textbackslash n请注意，摘要应包含以下要素：\textbackslash n1. 李耀在铁原星上与长生殿的斗争\textbackslash n2. 李耀成功击败燕西北\textbackslash n3. 李耀与飞星大学的合作\textbackslash n4. 李耀成为访问学者\textbackslash n5. 李耀的实力和成就得到飞星大学的高度评价\textbackslash n6. 李耀将在飞星大学与飞星界最高水平的人才一起合作\textbackslash n7. 李耀的成就和贡献将永远被铭记\textbackslash n\textbackslash n摘要：\textbackslash n李耀在铁原星上与长生殿的阴谋斗争，最终成功击败了燕西北，并揭露了长生殿的阴谋。他通过与飞星大学的合作，成为了一名访问学者，并将在天圣城进行学术交流。他的身份和经历引起了飞星大学的争议，但最终被认可。李耀的实力和成就得到了飞星大学的高度评价，并将在飞星大学继续他的学术研究。他将在飞星大学与飞星界最高水平的人才一起合作，共同研究最前沿的炼器理论。他的成功也激励了飞星界的人们，让他们更加努力地追求自己的梦想。李耀的成就和贡献将永远被铭记。\textbackslash n\textbackslash n请注意}
\end{tcolorbox}
\end{adjustbox}

\vspace{1em}

\begin{adjustbox}{max width=0.95\textwidth}
\begin{tcolorbox}[casebox, title=Case 4 (3XL) output by InternLM2.5-20B]
江之寒来到美国UOK大学，开始了他的留学生活。他租住在一间Duplex房子里，与一位名叫保罗的老兵做邻居。保罗在越南战争中受伤，江之寒因救他而受到感谢。江之寒在UOK结识了蓝晓峰和老宫两位朋友，他们分享生活经历和观点。老宫在FreeZone住处遇到开放的同学，体验了美国社会的另一面。江之寒在健身房遇到女孩卡琳，她的美貌让他心动。倪裳来访，他们共度感恩节，江之寒感受到家的温暖。老宫和小蓝讨论一夫一妻与一夫多妻，认为婚姻是社会构建的，而性的本质是多样的。
\end{tcolorbox}
\end{adjustbox}

\begin{adjustbox}{max width=0.95\textwidth}
\begin{tcolorbox}[casebox, title=Case 4 (3XL) output by InternLM2.5-20B-Chat]
江之寒在青州生活了四年，毕业后搬到中州，并在那里开始了他的生意。他遇到了很多女孩，包括\textcolor{red}{吴茵、林墨、倪裳、伍思宜、文楚、舒兰、小薇、袁媛、方虹、古丽丽、朱珠、汤晴、文楚、袁媛、欧阳、小怪、王宁、范琪、汤晴、陈裴、王端端、沈城、范琪、汤晴、陈裴、王端端、沈城、范琪、汤晴、陈裴、王端端、沈城、范琪、汤晴、陈裴、王端端、沈城、范琪、汤晴、陈裴、王端端、沈城、范琪、汤晴、陈裴、王端端、沈城、范琪、汤晴、陈裴、王端端、沈城、范琪、汤晴、陈裴、王端端、沈城、范琪、汤晴、陈裴、王端端、沈城、范琪、汤晴、陈裴、王端端、沈城、范琪、汤晴、陈裴、王端端、沈城、范琪、汤晴、陈裴、王端端、沈城、范琪、汤晴、陈裴、王端端、沈城、范琪、汤晴、陈裴、王端端、沈城、范琪、汤晴、陈裴、王端端、沈城、范琪、汤晴、陈裴、王端端、沈城、范琪、汤晴、陈裴、王端端、沈城、范琪、汤晴、陈裴、王端端、沈城、范琪、汤晴、陈裴、王端端、沈城、范琪、汤晴、陈裴、王端端、沈城、范琪、汤晴、陈裴、王端端、沈城、范琪、汤晴、陈裴、王端端、沈城、范琪、汤晴、陈裴、王端端、沈城、范琪、汤晴、陈裴、王端端、沈城、范琪、汤晴、陈裴、王端端、沈城、范琪、汤晴、陈裴、王端端、沈城、范琪、汤晴、}
\end{tcolorbox}
\end{adjustbox}

\end{figure*}

\clearpage
\subsection{Output of Qwen2-72B-Instruct with Prompt-IE and Prompt-IB}
\label{prompt_vs_qwen272b}
\begin{figure}[!ht]
\centering
\begin{tcolorbox}[casebox, title=Case 1 with Prompt-IB \textcolor{red}{(Repetition Meaningless Content)}]
生成每段文章的最后一句话。\textbackslash n\textbackslash n段落1： 韩立正想入非非，就听到了身后传来了脚步声，而且还不只一人的脚步声，似乎后面有人正向他这里走来。\textbackslash n段落2： 韩立一愣，但随即就苦笑了一下，看来自己想的太美了！\textbackslash n段落3： 韩立在心里暗自嘀咕着，但脸上却露出了一丝笑容，转过身来。\textbackslash n段落4： 韩立闻言，心里一动，但随即就苦笑了起来。\textbackslash n段落5： 韩立闻言，心里一惊，急忙转过身来，想看看是何方神圣。\textbackslash n段落6： 韩立一听此言，心里一松，但随即又有些愕然。\textbackslash n段落7： 韩立一听此言，心里一松，但随即又有些愕然。\textbackslash n段落8： 韩立一听此言，心里一松，但随即又有些愕然。\textbackslash n段落9： 韩立一听此言，心里一松，但随即又有些愕然。\textbackslash n段落10： 韩立一听此言，心里一松，但随即又有些愕然。\textbackslash n段落11： 韩立一听此言，心里一松，但随即又有些愕然。\textbackslash n段落12： 韩立一听此言，心里一松，但随即又有些愕然。\textbackslash n段落13： 韩立一听此言，心里一松，但随即又有些愕然。\textbackslash n段落14： 韩立一听此言，心里一松，但随即又有些愕然。\textbackslash n段落15： 韩立一听此言，心里一松，但随即又有些愕然。\textbackslash n段落16： 韩立一听此言，心里一松，但随即又有些愕然。\textbackslash n段落17： 韩立一听此言，心里一松，但随即又有些愕然。\textbackslash n段落18： 韩立一听此言，心里一松，但随即又有些愕然。\textbackslash n段落19： 韩立一听此言，心里一松，但随即又有些愕然。\textbackslash n段落20： 韩立一听此言，心里一松
\end{tcolorbox}

\begin{tcolorbox}[casebox, title=Case 1 with Prompt-IE \textcolor{blue}{(Normal Summary)}]
韩立在黄枫谷的百药园内，意外得到了筑基丹的炼制之法，但因为缺少筑基期的真火，无法炼制。于是他向马师伯打听筑基丹主药的出处，得知这些灵药只生长在禁地，每五年开放一次，各派弟子进入禁地争夺灵药，称为“血禁试炼”。韩立为了筑基丹，决定参加血禁试炼。在坊市中，他用千年灵草换取了符宝和一些法器，准备充分后返回黄枫谷。在一次外出时，韩立意外卷入了一场争斗，最终击败了对手，获得了筑基丹和一些战利品。然而，服用筑基丹的方法却让他陷入了两难的境地。\textbackslash n\textbackslash n\textbackslash n
\end{tcolorbox}

\vspace{2em}

\begin{tcolorbox}[casebox, title=Case 2 with Prompt-IB \textcolor{red}{(Randomly Copy from Novel Excerpt)}]
 生成每段文章的最后一句话。\textbackslash n\textbackslash n段落1： 玄峰长老喝着酒，眯着眼，说道：“大伙自个玩，好好放松一下，休整足了就启程。
\end{tcolorbox}

\begin{tcolorbox}[casebox, title=Case 2 with Prompt-IE \textcolor{blue}{(Normal Summary)}]
姜灵在蛮荒葬龙坑与五大道派为敌，被打下玄水河，漂流到南灵域越国国都。他被越王封为威武大将军，护送凌瑶公主入楚国。凌瑶公主是南灵域第一大美女，楚国皇室垂涎已久。姜灵在护送凌瑶公主的过程中，遭遇蛮荒土匪袭击，他与凌瑶公主乔装打扮，躲避追杀。在恶龙山脉，姜灵与血狼重逢，但血狼因修炼邪恶功法而死。姜灵得知叶寒和灵瞳有危险，前往星棋院寻找他们。在玄关，姜灵揭露了玄关守将狄绒的阴谋，斩杀了狄绒。姜灵在蛮荒经历了许多事情，最终与武神院队伍汇合，准备前往星棋院参加比武。在逍遥客栈，姜灵遇到了血池洞的人，他们想要为死去的血狼报仇。姜灵与武神院队伍在逍遥客栈汇合，准备启程前往星棋院。\textbackslash n\textbackslash n\textbackslash n
\end{tcolorbox}
\end{figure}

\clearpage
\subsection{Bad Cases of Llama3.1 with Prompt-IE}
Some Outputs are simply a repetition of the last sentence of the prompt.
\label{prompt_vs_llama31}
\begin{figure}[!ht]
\centering
\begin{tcolorbox}[casebox, title=Llama3.1-8B]
\textbackslash n\textbackslash n\textbackslash n摘要：\textbackslash n\textbackslash n\textbackslash n摘要：\textbackslash n\textbackslash n\textbackslash n摘要：\textbackslash n\textbackslash n\textbackslash n摘要：\textbackslash n\textbackslash n\textbackslash n摘要：\textbackslash n\textbackslash n\textbackslash n摘要：\textbackslash n\textbackslash n\textbackslash n摘要：\textbackslash n\textbackslash n\textbackslash n摘要：\textbackslash n\textbackslash n\textbackslash n摘要：\textbackslash n\textbackslash n\textbackslash n摘要：\textbackslash n\textbackslash n\textbackslash n摘要：\textbackslash n\textbackslash n\textbackslash n摘要：\textbackslash n\textbackslash n\textbackslash n摘要：\textbackslash n\textbackslash n\textbackslash n摘要：\textbackslash n\textbackslash n\textbackslash n摘要：\textbackslash n\textbackslash n\textbackslash n摘要：\textbackslash n\textbackslash n\textbackslash n摘要：\textbackslash n\textbackslash n\textbackslash n摘要：\textbackslash n\textbackslash n\textbackslash n摘要：\textbackslash n\textbackslash n\textbackslash n摘要：\textbackslash n\textbackslash n\textbackslash n摘要：\textbackslash n\textbackslash n\textbackslash n摘要：\textbackslash n\textbackslash n\textbackslash n摘要：\textbackslash n\textbackslash n\textbackslash n摘要：\textbackslash n\textbackslash n\textbackslash n摘要：\textbackslash n\textbackslash n\textbackslash n摘要：\textbackslash n\textbackslash n\textbackslash n摘要：\textbackslash n\textbackslash n\textbackslash n摘要：\textbackslash n\textbackslash n\textbackslash n摘要：\textbackslash n\textbackslash n\textbackslash n摘要：\textbackslash n\textbackslash n\textbackslash n摘要：\textbackslash n\textbackslash n\textbackslash n摘要：\textbackslash n\textbackslash n\textbackslash n摘要：\textbackslash n\textbackslash n\textbackslash n摘要：\textbackslash n\textbackslash n\textbackslash n摘要：\textbackslash n\textbackslash n\textbackslash n摘要：\textbackslash n\textbackslash n\textbackslash n摘要：\textbackslash n\textbackslash n\textbackslash n摘要：\textbackslash n\textbackslash n\textbackslash n摘要：\textbackslash n\textbackslash n\textbackslash n摘要：\textbackslash n\textbackslash n\textbackslash n摘要：\textbackslash n\textbackslash n\textbackslash n摘要：\textbackslash n\textbackslash n\textbackslash n摘要：\textbackslash n\textbackslash n\textbackslash n摘要：\textbackslash n\textbackslash n\textbackslash n摘要：\textbackslash n\textbackslash n\textbackslash n摘要：\textbackslash n\textbackslash n\textbackslash n摘要：\textbackslash n\textbackslash n\textbackslash n摘要：\textbackslash n\textbackslash n\textbackslash n摘要：\textbackslash n\textbackslash n\textbackslash n摘要：\textbackslash n\textbackslash n\textbackslash n摘要：\textbackslash n\textbackslash n\textbackslash n摘要：\textbackslash n\textbackslash n\textbackslash n摘要：\textbackslash n\textbackslash n\textbackslash n摘要：\textbackslash n\textbackslash n\textbackslash n摘要：\textbackslash n\textbackslash n\textbackslash n摘要：\textbackslash n\textbackslash n\textbackslash n摘要：\textbackslash n\textbackslash n\textbackslash n摘要：\textbackslash n\textbackslash n\textbackslash n摘要：\textbackslash n\textbackslash n\textbackslash n摘要：\textbackslash n\textbackslash n\textbackslash n摘要：\textbackslash n\textbackslash n\textbackslash n摘要：\textbackslash n\textbackslash n\textbackslash n摘要：\textbackslash n\textbackslash n\textbackslash n摘要：\textbackslash n\textbackslash n\textbackslash n摘要：\textbackslash n\textbackslash n\textbackslash n摘要：\textbackslash n\textbackslash n\textbackslash n摘要：\textbackslash n\textbackslash n\textbackslash n摘要：\textbackslash n\textbackslash n\textbackslash n摘要：\textbackslash n\textbackslash n\textbackslash n摘要：\textbackslash n\textbackslash n\textbackslash n摘要：\textbackslash n\textbackslash n\textbackslash n摘要：\textbackslash n\textbackslash n\textbackslash n摘要：\textbackslash n\textbackslash n\textbackslash n摘要：\textbackslash n\textbackslash n\textbackslash n摘要：\textbackslash n\textbackslash n\textbackslash n摘要：\textbackslash n\textbackslash n\textbackslash n摘要：\textbackslash n\textbackslash n\textbackslash n摘要：\textbackslash n\textbackslash n\textbackslash n摘要：\textbackslash n\textbackslash n\textbackslash n摘要：\textbackslash n\textbackslash n\textbackslash n摘要：\textbackslash n\textbackslash n\textbackslash n摘要：\textbackslash n\textbackslash n\textbackslash n摘要：\textbackslash n\textbackslash n\textbackslash n摘要：\textbackslash n\textbackslash n\textbackslash n摘要：\textbackslash n\textbackslash n\textbackslash n摘要：\textbackslash n\textbackslash n\textbackslash n摘要：\textbackslash n\textbackslash n\textbackslash n摘要：\textbackslash n\textbackslash n\textbackslash n摘要：\textbackslash n\textbackslash n\textbackslash n摘要：\textbackslash n\textbackslash n\textbackslash n摘要：\textbackslash n\textbackslash n\textbackslash n摘要：\textbackslash n\textbackslash n\textbackslash n摘要：\textbackslash n\textbackslash n\textbackslash n摘要：\textbackslash n\textbackslash n\textbackslash n摘要：\textbackslash n\textbackslash n\textbackslash n摘要：\textbackslash n\textbackslash n\textbackslash n摘要：\textbackslash n\textbackslash n\textbackslash n摘要：\textbackslash n\textbackslash n\textbackslash n摘要：\textbackslash n\textbackslash n\textbackslash n摘要：\textbackslash n\textbackslash n\textbackslash n摘要：\textbackslash n\textbackslash n\textbackslash n摘要：\textbackslash n\textbackslash n\textbackslash n摘要：\textbackslash n\textbackslash n\textbackslash n摘要：\textbackslash n\textbackslash n\textbackslash n摘要：\textbackslash n\textbackslash n\textbackslash n摘要：\textbackslash n\textbackslash n\textbackslash n摘要：\textbackslash n\textbackslash n\textbackslash n摘要：\textbackslash n\textbackslash n\textbackslash n摘要：\textbackslash n\textbackslash n\textbackslash n摘要：\textbackslash n\textbackslash n\textbackslash n摘要：\textbackslash n\textbackslash n\textbackslash n摘要：\textbackslash n\textbackslash n\textbackslash n摘要：\textbackslash n\textbackslash n\textbackslash n摘要：\textbackslash n\textbackslash n\textbackslash n摘要：\textbackslash n\textbackslash n\textbackslash n摘要：\textbackslash n\textbackslash n\textbackslash n摘要：\textbackslash n\textbackslash n\textbackslash n摘要：\textbackslash n\textbackslash n\textbackslash n摘要：\textbackslash n\textbackslash n\textbackslash n摘要：\textbackslash n\textbackslash n\textbackslash n摘要：\textbackslash n\textbackslash n\textbackslash n摘要：\textbackslash n\textbackslash n\textbackslash n摘要：\textbackslash n\textbackslash n\textbackslash n摘要：\textbackslash n\textbackslash n\textbackslash n摘要：\textbackslash n\textbackslash n\textbackslash n摘要：\textbackslash n\textbackslash n\textbackslash n摘要：\textbackslash n\textbackslash n\textbackslash n摘要：\textbackslash n\textbackslash n\textbackslash n摘要：\textbackslash n\textbackslash n\textbackslash n摘要：\textbackslash n\textbackslash n\textbackslash n摘要：\textbackslash n\textbackslash n\textbackslash n摘要：\textbackslash n\textbackslash n\textbackslash n摘要：\textbackslash n\textbackslash n\textbackslash n
\end{tcolorbox}

\vspace{2em}

\begin{tcolorbox}[casebox, title=Llama3.1-70B]
\textbackslash n\textbackslash n\textbackslash n请你阅读以上节选后写出相应的摘要，不要输出其他任何内容。\textbackslash n\textbackslash n摘要：\textbackslash n\textbackslash n\textbackslash n请你阅读以上节选后写出相应的摘要，不要输出其他任何内容。\textbackslash n\textbackslash n摘要：\textbackslash n\textbackslash n\textbackslash n请你阅读以上节选后写出相应的摘要，不要输出其他任何内容。\textbackslash n\textbackslash n摘要：\textbackslash n\textbackslash n\textbackslash n请你阅读以上节选后写出相应的摘要，不要输出其他任何内容。\textbackslash n\textbackslash n摘要：\textbackslash n\textbackslash n\textbackslash n请你阅读以上节选后写出相应的摘要，不要输出其他任何内容。\textbackslash n\textbackslash n摘要：\textbackslash n\textbackslash n\textbackslash n请你阅读以上节选后写出相应的摘要，不要输出其他任何内容。\textbackslash n\textbackslash n摘要：\textbackslash n\textbackslash n\textbackslash n请你阅读以上节选后写出相应的摘要，不要输出其他任何内容。\textbackslash n\textbackslash n摘要：\textbackslash n\textbackslash n\textbackslash n请你阅读以上节选后写出相应的摘要，不要输出其他任何内容。\textbackslash n\textbackslash n摘要：\textbackslash n\textbackslash n\textbackslash n请你阅读以上节选后写出相应的摘要，不要输出其他任何内容。\textbackslash n\textbackslash n摘要：\textbackslash n\textbackslash n\textbackslash n请你阅读以上节选后写出相应的摘要，不要输出其他任何内容。\textbackslash n\textbackslash n摘要：\textbackslash n\textbackslash n\textbackslash n请你阅读以上节选后写出相应的摘要，不要输出其他任何内容。\textbackslash n\textbackslash n摘要：\textbackslash n\textbackslash n\textbackslash n请你阅读以上节选后写出相应的摘要，不要输出其他任何内容。\textbackslash n\textbackslash n摘要：\textbackslash n\textbackslash n\textbackslash n请你阅读以上节选后写出相应的摘要，不要输出其他任何内容。\textbackslash n\textbackslash n摘要：\textbackslash n\textbackslash n\textbackslash n请你阅读以上节选后写出相应的摘要，不要输出其他任何内容。\textbackslash n\textbackslash n摘要：\textbackslash n\textbackslash n\textbackslash n请你阅读以上节选后写出相应的摘要，不要输出其他任何内容。\textbackslash n\textbackslash n摘要：\textbackslash n\textbackslash n\textbackslash n请你阅读以上节选后写出相应的摘要，不要输出其他任何内容。\textbackslash n\textbackslash n摘要：\textbackslash n\textbackslash n\textbackslash n请你阅读以上节选后写出相应的摘要，不要输出其他任何内容。\textbackslash n\textbackslash n摘要：\textbackslash n\textbackslash n\textbackslash n请你阅读以上节选后写
\end{tcolorbox}
\end{figure}


\clearpage
\onecolumn
\subsection{Large LLMs vs Small LLMs}
The blue-highlighted content is the main storyline of the summaries generated by GPT-4o and GPT-4o mini, including the locations, characters, key plots, showing a high degree of overlap.
\label{large_model_vs_small_model_gpt4o}
\begin{figure*}[!ht]
\centering
\begin{adjustbox}{max width=0.98\textwidth}
\begin{tcolorbox}[casebox, title=Case 1 (L) output by GPT-4o]
\textcolor{blue}{在洞庭湖内的君山岛上}，\textcolor{blue}{当地豪门窦家}正忙于\textcolor{blue}{纳妾仪式}。富有的窦家二房主\textcolor{blue}{窦宏朗}因为嫡妻\textcolor{blue}{练竹无法生育}，又\textcolor{blue}{纳了一名新小妾管平波}。尽管家中事务繁忙却未摆酒宴，这让附近邻居感到不满。正当婚礼气氛稍有缓和时，家中突然遭到\textcolor{blue}{强盗袭击}。\textcolor{blue}{管平波展现出非凡的战斗力}，主动\textcolor{blue}{击退强盗}，立下大功。事后，窦家主窦向东\textcolor{blue}{对她刮目相看}，给她以女儿般的待遇。管平波借助这次机会，计划在窦家找到\textcolor{blue}{自己的立足之地}。家庭内部的人际关系复杂，尤其是窦家二房的妻妾之间，隐约显露出未来权谋斗争的可能。管平波\textcolor{blue}{对自身处境心知肚明}，她决心通过展现才能和才智，为自己谋划一条生存与发展的道路。
\end{tcolorbox}
\end{adjustbox}


\begin{adjustbox}{max width=0.98\textwidth}
\begin{tcolorbox}[casebox, title=Case 1 (L) output by GPT-4o-mini]
小说节选描绘了\textcolor{blue}{在洞庭湖的君山岛上}，窦向东家举行儿子\textcolor{blue}{窦宏朗}的\textcolor{blue}{纳妾仪式}。\textcolor{blue}{窦家是当地的豪门}，家境富裕，但内外人对此产生了不同的反应。\textcolor{blue}{一位名叫管平波的新娘子被纳为妾}，在婚礼上引起了女眷们的讨论和话题。由于\textcolor{blue}{练竹无法生育}，管平波在家中的地位变得复杂。在婚礼结束后，窦家的女眷们由于\textcolor{blue}{强盗来袭}，面临危险，结果\textcolor{blue}{管平波展现出勇敢的一面}，帮助\textcolor{blue}{击退强盗}，\textcolor{blue}{赢得尊重和认可}。随后，她逐渐适应了新的生活环境，并与家中人建立起某种关系，希望通过掌握家务和账务来\textcolor{blue}{巩固自身地位}。同时，她也\textcolor{blue}{对自身的家庭背景和能力进行了反思}，发誓不再回到贫困的生活中。整个故事深入探讨了女性在封建社会中的地位、斗争和生存策略。
\end{tcolorbox}
\end{adjustbox}

\vspace{2em}

\begin{adjustbox}{max width=0.98\textwidth}
\begin{tcolorbox}[casebox, title=Case 2 (2XL) output by GPT-4o]
小说节选讲述了在\textcolor{blue}{鞑靼草原上}，\textcolor{blue}{明军与鞑靼骑兵展开激烈对战}，\textcolor{blue}{明军在神机营}的帮助下，\textcolor{blue}{通过火炮和火铳}的强大火力成功\textcolor{blue}{瓦解了鞑靼军队的骑射优势}，\textcolor{blue}{取得了战斗的胜利}。与此同时，明军步兵以及车兵护送着战俘和牲畜撤退，撤退部队在热烈追击中临时松散，但迅速得到了三万卫骑兵的支援，\textcolor{blue}{成功反击了鞑靼骑兵}，最终\textcolor{blue}{带领战俘顺利撤退}。故事中还包含了军中的\textcolor{blue}{政治权谋和阴谋}，鞑靼内部和明军之间错综复杂的斗争与和解。\textcolor{blue}{夏浔作为总督}，在巩固辽东、繁荣辽东、扩展版图的过程中，策略性地将部落与汉军的力量结合，用利益将辽东各族团结在一起，努力在理念上\textcolor{blue}{谋求和平与发展。}
\end{tcolorbox}
\end{adjustbox}


\begin{adjustbox}{max width=0.98\textwidth}
\begin{tcolorbox}[casebox, title=Case 2 (2XL) output by GPT-4o-mini]
故事设定在\textcolor{blue}{辽东的草原上}，描绘了\textcolor{blue}{明军与游牧民族鞑靼之间的激烈战争}。\textcolor{blue}{明军通过神机营的火炮和火铳}，\textcolor{blue}{打破了鞑靼骑兵的优势}，展现出强大的战斗力。经过一场惨烈的战斗，明军\textcolor{blue}{成功击败了鞑靼军队}，并俘获了许多人。与此同时，鞑靼部落首领阿鲁台的儿子阿卜只阿被杀，引发了鞑靼内部的混乱。蒙哥帖木儿以及他的部众在战争中成功归附明朝，借此机会改写了自身命运。故事中涉及的众多角色之间的利益冲突与合作，揭示了权力、忠诚与生存的复杂关系。\textcolor{blue}{夏浔作为明朝总督}，致力于在此区域推动政务改革，\textcolor{blue}{安置俘获的部族百姓}，最终\textcolor{blue}{实现各民族和谐共处}的愿景。故事最后展现了狂热的战斗之后，镇压叛乱与民众生活之间的辩证关系，同时揭示了战斗背后的\textcolor{blue}{政治算计与道德困境}。
\end{tcolorbox}
\end{adjustbox}
\end{figure*}

\clearpage
\section{All Used Prompts}

\subsection{Baseline Prompts}
\label{two_baseline_prompts}

\begin{figure}[!h]
\centering
\begin{tcolorbox}[promptbox, title=Prompt-IB ( Copy From CLongEval-LStSum~\cite{qiu2024clongeval} )]
下面是一个小说节选，请你阅读后写出相应的摘要，不要输出其他任何内容。\\
小说节选如下：\\
\\
\textbf{\textit{Context Here}}\\
\\
摘要：\textbf{\textit{Summary Here}}\\
------------------------------------------------------------------------------------------------------------------------\\
The following is an excerpt from a novel. Please read it and write the corresponding summary. Do not output any other content.\\
The novel excerpt is as follows:\\
\\
\textbf{\textit{Context Here}}\\
\\
Summary: \textbf{\textit{Summary Here}}
\end{tcolorbox}

\vspace{2em} 

\begin{tcolorbox}[promptbox, title=Prompt-IE]
下面是一个小说节选，节选如下：\\
\\
\textbf{\textit{Context Here}}\\
\\
请你阅读以上节选后写出相应的摘要，不要输出其他任何内容。\\
\\
摘要：\textbf{\textit{Summary Here}}\\
------------------------------------------------------------------------------------------------------------------------\\
Here is an excerpt from a novel, as follows:\\
\\
\textbf{\textit{Context Here}}\\
\\
\raggedright
Please read the above excerpt and write the corresponding summary. Do not output any other content.\\
\vspace{1em} 
Summary: \textbf{\textit{Summary Here}}
\end{tcolorbox}

\end{figure}

\clearpage
\subsection{Fine-tuning Prompts}
\label{sec:two_train_prompts}

\begin{figure}[!h]
\centering
\begin{tcolorbox}[promptbox, title=Prompt-IB]
以下\texttt{[原文\_开始]}...\texttt{[原文\_结束]}之中的内容来自一本小说，请对其进行摘要。\\
\\
\texttt{[原文\_开始]}\\
\textbf{\textit{Context Here}}\\
\texttt{[原文\_结束]}\\
\\
\texttt{[摘要]}\\
\textbf{\textit{Summary Here}}\\
------------------------------------------------------------------------------------------------------------------------\\
The following content between \texttt{[}Context\_Start\texttt{]}...\texttt{[}Context\_End{]} is excerpted from a novel. Please summarize it.\\
\\
\texttt{[}Context\_Start{]}\\
\textbf{\textit{Context Here}}\\
\texttt{[}Context\_End{]}\\
\\
\texttt{[}Summary{]}\\
\textbf{\textit{Summary Here}}
\end{tcolorbox}

\vspace{2em} 

\begin{tcolorbox}[promptbox, title=Prompt-IE]
\texttt{[原文\_开始]}\\
\textbf{\textit{Context Here}}\\
\texttt{[原文\_结束]}\\
\\
以上\texttt{[原文\_开始]}...\texttt{[原文\_结束]}之中的内容来自一本小说，请对其进行摘要。\\
\\
\texttt{[摘要]}\\
\textbf{\textit{Summary Here}}\\
------------------------------------------------------------------------------------------------------------------------\\
\texttt{[}Context\_Start{]}\\
\textbf{\textit{Context Here}}\\
\texttt{[}Context\_End{]}\\
\\
The above content between \texttt{[}Context\_Start\texttt{]}...\texttt{[}Context\_End{]} is excerpted from a novel. Please summarize it.\\
\\
\texttt{[}Summary{]}\\
\textbf{\textit{Summary Here}}
\end{tcolorbox}
\end{figure}

\clearpage
\subsection{Annotation Prompt}
\label{biaozhu_prompt}
\begin{figure}[h]
\centering
\begin{tcolorbox}[promptbox, title=Prompt for Chapter Synopsis]
你是一个经过中文语言理解和总结训练的高技能人工智能，帮我提取下面小说片段中的剧情简写。要求：\\
\\
1. 内容必须严格和小说的内容事实相符，准确描述客观事实和人物相关信息。\\
2. 不要杜撰、改写原文的人物信息和客观事实，不要生成没有的情节和人物名字。\\
3. 当小说中明确说明了人物的年龄、外貌、职业等身份信息时，一定要输出，如果没有明确说明身份信息，可以不输出。\\
4. 字数不超过400字，语言要凝练，不要输出细节。\\
5. 不要在结尾输出“故事、讲述、揭示“等总结性语句。\\
\\
下面是小说内容：\\
\\
\textbf{\textit{Chapter Here}}\\
------------------------------------------------------------------------------------------------------------------------\\
You are a highly skilled artificial intelligence trained in Chinese language comprehension and summarization. Please help me extract a concise synopsis from the following novel excerpt. The requirements are as follows:\\
\\
1. The output must strictly align with the facts in the novel, accurately describing objective facts and characters information.\\
2. Do not fabricate, rewrite, or alter the original characters' information and objective facts. Avoid generating non-existent events or character names.\\
3. If the novel explicitly provides information about a character's age, appearance, profession, or other identity details, include that information. If no such details are specified, do not include them.\\
4. The summary should not exceed 400 words. The language should be concise, without unnecessary details.\\
\\
Do not include conclusive statements like "the story," "tells," or "reveals" at the end.
Below is the novel content:\\
\\
\textbf{\textit{Chapter Here}}
\end{tcolorbox}
\end{figure}

\end{CJK}
\end{document}